\documentclass[11pt]{article}
\usepackage{acl}

\usepackage{times}
\usepackage{latexsym}
\usepackage[T1]{fontenc}
\usepackage[utf8]{inputenc}
\usepackage{microtype}
\usepackage{inconsolata}

\usepackage{booktabs}
\usepackage{graphicx}
\usepackage{subcaption}
\usepackage{amsmath}
\usepackage{tabularx}
\usepackage{hyperref}
\usepackage{comment}
\usepackage{siunitx}
\usepackage{xspace}
\usepackage{tikz}
\usetikzlibrary{backgrounds,patterns.meta}
\usetikzlibrary{decorations.pathreplacing}
\usetikzlibrary{arrows.meta,positioning,fit,matrix,calc,shapes}
\usepackage{threeparttable}
\usepackage{adjustbox}
\usepackage{multirow}
\usepackage{float}


\usepackage{longtable}
\usepackage{pdflscape}
\usepackage{xurl}
\newcommand{\feat}[1]{\nolinkurl{#1}}

\newcommand{\Ta}{\ensuremath{T_{\!A}}\xspace}
\newcommand{\Tt}{\ensuremath{T_{\!T}}\xspace}
\newcommand{\Ti}{\ensuremath{T_{\!I}}\xspace}

\newcommand{\Ocat}{\ensuremath{\mathcal{O}}\xspace}

\newcommand{\TOA}{\ensuremath{\mathcal{TOA}}\xspace}

\newcommand{\Oold}{\ensuremath{\mathcal{O}_{\!\text{ old}}}\xspace}
\newcommand{\Orec}{\ensuremath{\mathcal{O}_{\!\text{ recent}}}\xspace}

\newcommand{\TOAfirst}{\ensuremath{\mathcal{TOA}_{\!\text{ first}}}\xspace}
\newcommand{\TOAlate}{\ensuremath{\mathcal{TOA}_{\!\text{ late}}}\xspace}
\newcommand{\TODlate}{\ensuremath{\mathcal{TOD}_{\!\text{ late}}}\xspace}

\title{Predicting Emerging Topics from Outliers:\\
A Prospective Study of Weak Signals in Embedding Space}

\author{
  Evangelia Zve\textsuperscript{1,2}\thanks{Corresponding author.}, Gauvain Bourgne\textsuperscript{1},
  Jean-Gabriel Ganascia\textsuperscript{1} \\[0.3em]
  \textsuperscript{1}LIP6, Sorbonne Université, CNRS, \texttt{\{name.surname\}@lip6.fr} \\
  [0.3em]
  \textsuperscript{2}Infopro Digital
}
\begin{document}
\maketitle

\begin{abstract}Some documents that embedding-based topic models initially classify as noise later become founding members of emerging topics. At publication time, however, they appear as scattered points in embedding space and are difficult to distinguish from ordinary noise without the benefit of hindsight. We study whether such \emph{anticipatory outliers} can be predicted prospectively, using only information available when a document first appears. We derive labels from the subsequent trajectories of outlier documents, distinguishing those that anticipate new topics from those that reinforce existing topics or remain isolated, and estimate label confidence through agreement across multiple embedding models. On two French news corpora, anticipatory outliers prove predictable at publication time. Under cross-validation, \(F_1\) rises from about 0.77 over the full eligible population to above 0.90 on high-consensus subsets, and remains at 0.76--0.80 under a strictly chronological evaluation. Predictive performance is driven mainly by geometric features capturing each outlier's position in embedding space.\end{abstract} 

\section{Introduction}
\label{sec:introduction}

In document streams such as news or scientific literature, the challenge is not only to summarize topics that are already visible, but to identify topics as they begin to emerge \cite{allan2002topic,boutaleb2024bertrend,ebadi2026wisdom}. Despite recent advances in language models and representation learning, early topic detection remains difficult for embedding-based topic models.

\begin{figure}[t]
\centering
\resizebox{\linewidth}{!}{
\begin{tikzpicture}[
    x=1cm,y=1cm,
    >=Latex,
    every node/.style={font=\footnotesize},
    keylabel/.style={font=\footnotesize},
    keytime/.style={font=\footnotesize},
    oldtopic/.style={draw=black!80, dashed, thick},
    newtopic/.style={draw=black!85, thick, fill=black!4},
    pasttopic/.style={draw=gray!55, dashed, thick, fill=gray!8},
    futuretopic/.style={draw=gray!70, dashed, thick},
    art/.style={circle, fill=black, inner sep=0.9pt},
    pastart/.style={circle, fill=gray!55, inner sep=0.9pt},
    futureart/.style={circle, fill=gray!65, inner sep=0.9pt},
    focusdot/.style={
    circle,
    fill=blue!55!black,
    draw=black,
    line width=0.25pt,
    inner sep=1.2pt
    },
    signal/.style={
        draw=blue!55!black,
        very thick,
        densely dashed,
        ->,
        line cap=round
    },
    faint/.style={font=\footnotesize\itshape, text=black!65},
    guide/.style={densely dashed, gray!65},
    strongguide/.style={
        dash pattern=on 2.2pt off 1pt,
        black!70,
        line width=1.05pt
    },
    timelabel/.style={fill=white, inner sep=1pt}
]

\def\xstart{0.4}
\def\xPastEnd{2.8}
\def\xTrainEnd{6.9}
\def\xFutureStart{10.9}
\def\xend{13.8}

\def\cPast{1.6}
\def\cTA{4.85}
\def\cTTI{8.9}
\def\cFuture{12.35}

\def\yTrain{-1.05}
\def\yTrainCapTop{-0.93}
\def\yTrainCapBot{-1.17}
\def\yTrainLabel{-1.35}

\draw[thick,->] (\xstart,0) -- (\xend,0) node[right] {time};

\draw[guide] (\xPastEnd,\yTrain) -- (\xPastEnd,2.90);
\draw[guide] (\xTrainEnd,\yTrain) -- (\xTrainEnd,2.90);

\draw[strongguide] (\xFutureStart,\yTrain) -- (\xFutureStart,2.90);
\draw[black!70, line width=0.8pt]
    (\xFutureStart-0.08,2.90) -- (\xFutureStart+0.08,2.90);
\draw[black!70, line width=0.8pt]
    (\xFutureStart-0.08,\yTrain) -- (\xFutureStart+0.08,\yTrain);

\draw[thick] (\cPast,0.12) -- (\cPast,-0.12);
\draw[thick] (\cTA,0.16) -- (\cTA,-0.16);
\draw[thick] (\cTTI,0.16) -- (\cTTI,-0.16);
\draw[thick] (\cFuture,0.12) -- (\cFuture,-0.12);

\node[timelabel] at (\cPast,-0.30) {past};
\node[keytime,timelabel] at (\cTA,-0.30) {$T_A$};
\node[keytime,timelabel] at (\cTTI,-0.30) {$T_T = T_I$};
\node[timelabel] at (\cFuture,-0.30) {future};

\node[keylabel,align=center] at (\cTA,2.66) {publication-time\\snapshot};
\node[keylabel,align=center] at (\cTTI,2.66) {new topic\\forms};
\node[keylabel,align=center] at (\cFuture,2.66) {deployment-time\\snapshot};

\draw[pasttopic] (\cPast,1.25) circle (0.34);
\node[pastart] at (\cPast-0.14,1.37) {};
\node[pastart] at (\cPast+0.05,1.40) {};
\node[pastart] at (\cPast+0.13,1.21) {};
\node[pastart] at (\cPast-0.06,1.10) {};

\draw[oldtopic] (\cTA-0.55,1.25) circle (0.36);
\node[art] at (\cTA-0.70,1.38) {};
\node[art] at (\cTA-0.48,1.42) {};
\node[art] at (\cTA-0.38,1.22) {};
\node[art] at (\cTA-0.62,1.10) {};

\draw[oldtopic] (\cTA+0.35,1.55) circle (0.46);
\node[art] at (\cTA+0.15,1.72) {};
\node[art] at (\cTA+0.40,1.77) {};
\node[art] at (\cTA+0.54,1.55) {};
\node[art] at (\cTA+0.21,1.36) {};
\node[art] at (\cTA+0.44,1.28) {};

\node[focusdot] (iA) at (\cTA+1.25,1.33) {};
\node[keylabel,right=1.5pt] at (iA) {$i$};
\node[faint,below] at (\cTA+1.00,0.88) {outlier};

\draw[oldtopic] (\cTTI-0.80,1.25) circle (0.36);
\node[art] at (\cTTI-0.95,1.38) {};
\node[art] at (\cTTI-0.73,1.42) {};
\node[art] at (\cTTI-0.63,1.22) {};
\node[art] at (\cTTI-0.87,1.10) {};

\draw[oldtopic] (\cTTI+0.00,1.55) circle (0.46);
\node[art] at (\cTTI-0.20,1.72) {};
\node[art] at (\cTTI+0.05,1.77) {};
\node[art] at (\cTTI+0.19,1.55) {};
\node[art] at (\cTTI-0.14,1.36) {};
\node[art] at (\cTTI+0.09,1.28) {};
\node[art] at (\cTTI-0.28,1.55) {};
\node[art] at (\cTTI+0.00,1.55) {};

\draw[newtopic] (\cTTI+1.00,1.35) circle (0.42);
\node[art] at (\cTTI+0.83,1.49) {};
\node[art] at (\cTTI+1.02,1.54) {};
\node[art] at (\cTTI+1.18,1.30) {};
\node[art] at (\cTTI+0.91,1.18) {};

\node[focusdot] (iB) at (\cTTI+1.08,1.35) {};
\node[faint,below] at (\cTTI+1.00,0.88) {new topic};

\draw[signal]
    (iA) .. controls (\cTA+1.75,2.12) and (\cTTI-0.20,2.12) .. (iB);

\draw[futuretopic] (\cFuture-0.80,1.25) circle (0.36);
\node[futureart] at (\cFuture-0.95,1.38) {};
\node[futureart] at (\cFuture-0.73,1.42) {};
\node[futureart] at (\cFuture-0.63,1.22) {};
\node[futureart] at (\cFuture-0.87,1.10) {};

\draw[futuretopic] (\cFuture+0.00,1.55) circle (0.46);
\node[futureart] at (\cFuture-0.20,1.72) {};
\node[futureart] at (\cFuture+0.05,1.77) {};
\node[futureart] at (\cFuture+0.19,1.55) {};
\node[futureart] at (\cFuture-0.14,1.36) {};
\node[futureart] at (\cFuture+0.09,1.28) {};
\node[futureart] at (\cFuture-0.28,1.55) {};
\node[futureart] at (\cFuture+0.00,1.55) {};

\draw[futuretopic] (\cFuture+0.90,1.35) circle (0.42);
\node[futureart] at (\cFuture+0.73,1.49) {};
\node[futureart] at (\cFuture+0.96,1.52) {};
\node[futureart] at (\cFuture+1.07,1.32) {};
\node[futureart] at (\cFuture+0.81,1.18) {};
\node[futureart] at (\cFuture+0.90,1.35) {};

\node[futureart] at (\cFuture+1.10,1.48) {};

\node[focusdot] (j) at (\cFuture+0.55,2.10) {};
\node[keylabel,right=1.5pt] at (j) {$j$};
\node[faint,anchor=west] at (\cFuture+0.88,1.98) {new outlier};

\draw[thick] (\xPastEnd,\yTrain) -- (\xFutureStart,\yTrain);
\draw[thick] (\xPastEnd,\yTrainCapTop) -- (\xPastEnd,\yTrainCapBot);
\draw[thick] (\xFutureStart,\yTrainCapTop) -- (\xFutureStart,\yTrainCapBot);

\node[keylabel,align=center] at (\cTA,-0.82) {features};
\node[keylabel,align=center] at (\cTTI,-0.82) {labels};

\node[faint] at ({(\xPastEnd+\xFutureStart)/2},\yTrainLabel)
{retrospective supervision for training};

\draw[->,thick] (\xFutureStart,\yTrain) -- (\cFuture+1.45,\yTrain);
\node[faint] at ({(\xFutureStart+\cFuture+1.10)/2},\yTrainLabel)
{prospective prediction};

\end{tikzpicture}
}
\caption{Prospective prediction under retrospective supervision. Labels are reconstructed from later trajectories. Predictors are computed at publication time $T_A$. Article $i$ illustrates an anticipatory outlier.}
\label{fig:prediction_setup}
\end{figure}
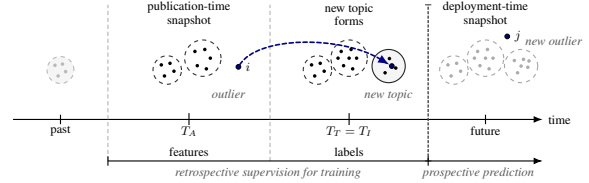

Methods such as BERTopic \cite{grootendorst2022bertopic} place documents in a semantic space and cluster them, often using density-based algorithms such as HDBSCAN \cite{mcinnes2017hdbscan}. Documents outside dense regions are usually treated as noise and excluded. Yet these outliers do not all share the same fate \cite{zve2025outliers}. Some remain isolated, some later join topics that already existed, and some precede topics that had not yet formed.

Building on the temporal taxonomy of document trajectories introduced by \citet{zve-etal-2026-noise}, we focus on documents that appear before the topics they later join. Retrospectively, this anticipatory status is observable once the future topic has formed. Prospectively, however, the same document is still only an outlier in the current embedding space. This gap raises a concrete prediction problem. Can publication-time evidence distinguish ordinary outliers from documents that will later become part of a topic that has not yet formed?

To address this, we reconstruct article trajectories from cumulative daily snapshots of a news stream. On each day, articles observed so far are clustered in embedding space, and the resulting clusters are aligned across consecutive snapshots, to track continuing topics and detect newly formed ones. This reconstruction allows us to determine whether publication-time outliers remain isolated, later join existing topics, or anticipate topics that form only after the article appears.

We formulate anticipatory outlier detection as a supervised binary classification task in which retrospective topic trajectories provide labels, while predictors are restricted to information available when each document first appears. Retrospective topic formation provides the supervision needed to train a prospective model.

We make three contributions. First, we frame anticipatory outlier detection as a prospective supervised article-level prediction task, with labels reconstructed from retrospective topic trajectories across multiple embedding models and confidence estimated through inter-model agreement. Second, we propose and discuss three complementary families of publication-time predictors: geometric position, textual content, and early social circulation of outlier articles. Third, we show that anticipatory outliers are predictable at publication time, both under cross-validation and under a strictly chronological evaluation, and identify the features that drive these predictions.

\section{Related Work}
\label{sec:related-work}

Weak-signal analysis studies sparse, ambiguous observations whose significance becomes clear only in retrospect \cite{hiltunen2008future,ansoff1975weak}. Related NLP work on Topic Detection and Tracking \cite{allan2002topic}, first-story detection \cite{petrovic2010streaming}, burst detection \cite{kleinberg2002bursty}, and social-stream event detection \cite{becker2011beyond} seeks early evidence of emerging developments, typically at the event, story, burst, or topic level. In contrast, work on topic-model outliers treats initially unclustered articles as the unit of analysis and asks whether they become early members of emerging topics \cite{zve2025outliers,zve-etal-2026-noise}.

Our task is also related to novelty and outlier detection, which identify deviations from regularities \cite{pimentel2014review,aggarwal2017outlier}. Here, outlier status is only the starting point. Rather than detecting outliers per se, we ask which publication-time outliers later become precursors of new topics.

Our work also relates to temporal and embedding-based topic modeling. Classical dynamic topic models track latent topics from word co-occurrence patterns over time \cite{blei2006dynamic,wang2006topics}. Neural and embedding-based topic models instead represent documents in dense semantic spaces, improving topic discovery through contextual representations \cite{dieng2020topic,grootendorst2022bertopic}. Recent emerging-topic systems track embedding-space topical structure by aligning clusters across time windows and monitoring nascent topics \cite{christophe2021monitoring,boutaleb2024bertrend}. These methods are well suited to tracking topics once repeated evidence has accumulated \cite{ebadi2026wisdom}. We instead focus on an earlier stage, when articles remain unclustered and their future topical role is unclear, asking whether these trajectories can be predicted at publication time.

\section{Trajectory-Based Supervision}
\label{sec:labeling}

We derive supervised labels from the later trajectories of outlier articles, following the taxonomy of \citet{zve-etal-2026-noise}. On each day, all articles published so far are clustered in embedding space, which assigns each article either to a topic or to the outlier set. Aligning topics across consecutive days then tells us whether each topic is continuing or newly created. An article that was an outlier at publication time can therefore end up in one of three situations: it joins a topic that already existed when it appeared, it remains an outlier, or it joins a topic that only formed after its publication. The label is positive in this last, anticipatory case. We run this procedure separately for each embedding model and combine the results in Section~\ref{sec:labeling:consensus}. Figure~\ref{fig:prediction_setup} illustrates the setup with one anticipatory outlier.

\subsection{Cumulative Clustering Setting}
\label{sec:labeling:reconstruction}

For each day \(t\), the snapshot contains all articles observed up to and including \(t\). Article texts are embedded, projected with UMAP \cite{mcinnes2018umap}, and clustered using a density-based method. Such methods, including HDBSCAN \cite{mcinnes2017hdbscan} and OPTICS \cite{ankerst1999optics}, can identify low-density observations that do not belong to any cluster. These observations form the candidate outlier cases considered below.

Topic continuity is recovered by aligning clusters in consecutive daily snapshots. Each cluster is represented by the centroid of its article embeddings in the reduced space. The one-to-one matching problem between clusters at \(t-1\) and \(t\) is solved using the Hungarian algorithm \cite{kuhn1955hungarian}, with cosine distance between centroids as the matching cost. A cluster at time \(t\) inherits the identity of a previous topic if its best match lies below the alignment threshold \(\theta_{\mathrm{align}}\); otherwise, it is treated as a new topic. This aligned sequence provides the temporal reference used to assign outlier trajectories.

\subsection{Trajectories and Prediction Target}
\label{sec:labeling:target}

Using the notation of \citet{zve-etal-2026-noise}, each article trajectory is characterized by three temporal variables: its publication time \(\Ta\), the creation time \(\Tt\) of the topic it eventually joins, if any, and its first integration time \(\Ti\) into that topic. The relative ordering of these events determines the trajectory category and defines the classification target.

The \emph{positive class} consists of \emph{anticipatory outliers}: articles that are unassigned at publication time and later join a topic that did not yet exist when they appeared, i.e., \(\Ta < \Tt \le \Ti\). This includes \(\TOAfirst\), where integration occurs when the topic is created, \(\Ta < \Tt = \Ti\), and \(\TOAlate\), where the topic is created after publication but the article joins it only later, \(\Ta < \Tt < \Ti\).

The \emph{negative class} is the complement of this target within the set of articles that are outliers at \Ta. It includes articles that later join a topic that already existed when they appeared, denoted by \(\TODlate\), and articles that remain unassigned throughout the observation window, denoted by \(\Ocat = \Orec \cup \Oold\). The supervised task is therefore the binary distinction \(\TOA = \TOAfirst \cup \TOAlate\) versus \(\neg \TOA = \TODlate \cup \Ocat\), restricted to articles that are outliers at publication time.

\subsection{Consensus-Based Filtering and Labeling}
\label{sec:labeling:consensus}

A central difficulty is that trajectory assignment is model-dependent. The embedding space, density structure, and cross-time topic alignment all depend on the upstream embedding model. We therefore do not treat any single model-specific trajectory assignment as definitive supervision. Instead, we run the trajectory-labeling pipeline separately for each embedding model and use inter-model agreement in two ways: to filter reliable publication-time outliers and to assign anticipatory versus non-anticipatory labels. This is aligned with previous work showing that agreement across imperfect labeling sources can serve as a proxy for label reliability \cite{snow2008cheap,ratner2017snorkel,strehl2002cluster}.

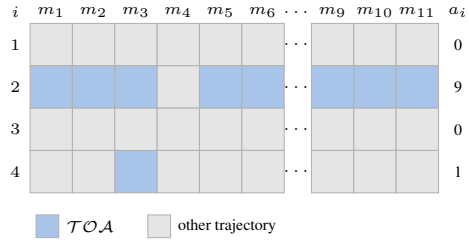
\begin{figure}[t]
\centering
\begin{adjustbox}{max width=\columnwidth}
\begin{tikzpicture}[
   font=\scriptsize,
   cell/.style={draw=gray!60, minimum size=5.6mm, inner sep=0pt, font=\tiny},
   lab/.style={font=\tiny\itshape, align=center}]
\definecolor{toaC}{RGB}{168,196,232}
\definecolor{notC}{RGB}{229,229,229}
\matrix (A) [matrix of nodes, nodes={cell, anchor=center},
   column sep=-\pgflinewidth, row sep=-\pgflinewidth,
   row 1/.style={nodes={draw=none, font=\tiny\itshape, minimum size=3.2mm}},
   column 1/.style={nodes={draw=none, font=\tiny\itshape, minimum size=4mm}},
   column 8/.style={nodes={draw=none, font=\tiny, minimum size=3.4mm}},
   column 12/.style={nodes={draw=none, minimum size=3mm}},
   column 13/.style={nodes={draw=none, font=\tiny, minimum size=5.6mm}}]
{
  $i$ & $m_1$ & $m_2$ & $m_3$ & $m_4$ & $m_5$ & $m_6$ & $\cdots$ & $m_9$ & $m_{10}$ & $m_{11}$ & & $a_i$ \\
 $1$ & |[fill=notC]| & |[fill=notC]| & |[fill=notC]| & |[fill=notC]| & |[fill=notC]|
       & |[fill=notC]| & $\cdots$ & |[fill=notC]| & |[fill=notC]| & |[fill=notC]| & & 0 \\
 $2$ & |[fill=toaC]| & |[fill=toaC]| & |[fill=toaC]| & |[fill=notC]| & |[fill=toaC]|
       & |[fill=toaC]| & $\cdots$ & |[fill=toaC]| & |[fill=toaC]| & |[fill=toaC]| & & 9 \\
 $3$ & |[fill=notC]| & |[fill=notC]| & |[fill=notC]| & |[fill=notC]| & |[fill=notC]|
       & |[fill=notC]| & $\cdots$ & |[fill=notC]| & |[fill=notC]| & |[fill=notC]| & & 0 \\
 $4$ & |[fill=notC]| & |[fill=notC]| & |[fill=toaC]| & |[fill=notC]| & |[fill=notC]|
       & |[fill=notC]| & $\cdots$ & |[fill=notC]| & |[fill=notC]| & |[fill=notC]| & & 1 \\
};
\node[cell, fill=toaC, anchor=west, minimum size=3mm] (lg1)
  at ([shift={(6mm,-3mm)}]A.south west) {};
\node[font=\tiny, anchor=west, inner sep=1pt] (lg1t) at (lg1.east) {~\TOA};
\node[cell, fill=notC, anchor=west, minimum size=3mm] (lg2)
  at ([xshift=4mm]lg1t.east) {};
\node[font=\tiny, anchor=west, inner sep=1pt] at (lg2.east) {~other trajectory};
\end{tikzpicture}
\end{adjustbox}
\caption{Consensus labeling example. Rows are articles \(i=1,\dots,4\), columns \(m_1\)--\(m_{11}\) are the embedding models in \(\mathcal{M}\), and \(a_i\) is the number of models assigning the \(\TOA\) trajectory to article \(i\) (shaded cells). The four articles have already passed the eligibility filter \(o_i\ge k_o\). Under \((k,k,0)\), article 2 is positive for \(k\le 9\), article 4 only for \(k=1\), and articles 1 and 3 are negative for every \(k\).}
\label{fig:agreement-matrix}
\end{figure}

Let \(\mathcal{M}\) denote the set of embedding models. For each article \(i\), let \(o_i\) be the number of models that classify the article as an outlier at publication time, and let \(a_i\) be the number of models that assign it to an anticipatory trajectory. We define labels using three consensus thresholds \((k_o,k_a,k_n)\), an outlier-eligibility threshold \(k_o\), a positive-label threshold \(k_a\), and a negative-label threshold \(k_n\).
\[
\small
y_i =
\begin{cases}
1, & o_i \ge k_o \ \text{and}\ a_i \ge k_a,\\
0, & o_i \ge k_o \ \text{and}\ a_i \le k_n,\\
\text{uncertain}, & \text{otherwise.}
\end{cases}
\]
The condition \(o_i \ge k_o\) defines the eligible population: articles reliably identified as publication-time outliers. Within this population, positives require at least \(k_a\) anticipatory votes, while negatives require at most \(k_n\) such votes. Articles satisfying neither rule are treated as \emph{uncertain} and excluded from training and evaluation. Figure~\ref{fig:agreement-matrix} illustrates the rule on four articles.

In our experiments, we use the conservative setting \((k_o,k_a,k_n)=(k,k,0)\). Thus, both outlier status and positive labels must be supported by at least \(k\) models, whereas negatives must receive no anticipatory vote. Varying \(k\) quantifies the coverage--confidence tradeoff. Larger values retain fewer articles but provide higher-confidence supervision. More general asymmetric rules are possible, but varying the three thresholds independently introduces additional calibration choices; we leave their systematic investigation to future work.

\section{What Predicts Anticipatory Outliers?}
\label{sec:predictors}

If anticipatory outliers are meaningful early signals rather than residual noise, they should already leave detectable traces at \(\Ta\). We therefore propose and discuss three complementary families of \(\Ta\)-based predictors, each capturing a different article-level source of evidence about later topic formation.

Geometric features locate the article in the publication-time embedding space, measuring whether it sits at the margins of established topic regions or in sparse residual areas where new structure may later form. Textual features characterize the article's content, tone, and form, testing whether anticipatory trajectories are associated with distinctive framing, entity anchoring, or stylistic cues. Social features capture early circulation on X, testing whether articles that later join emerging topics already leave observable traces in audience reach or audience-overlap structure. All predictors use only
\Ta features, while labels are derived from later trajectories to prevent future-topic leakage.\footnote{Full glossary of features is available in Appendix~\ref{app:glossary}.}

\subsection{Geometric Features}
\label{sec:predictor_families_geometry}

Geometric features measure where an outlier lies in the UMAP-reduced embedding space of the cumulative topic snapshot at \(\Ta\). Let \(x_i\) denote the reduced embedding of article \(i\), and let \(C_{\Ta}\) denote the set of topic clusters present at \(\Ta\).

\begin{figure}[t]
\centering
\resizebox{\columnwidth}{!}{
\begin{tikzpicture}[
    x=1cm,y=1cm,
    >=Latex,
    every node/.style={font=\footnotesize},
    title/.style={font=\footnotesize\bfseries},
    topic/.style={draw=black!75, dashed, line width=0.8pt},
    pt/.style={circle, fill=black, draw=black, inner sep=0.75pt},
    centroidmark/.style={font=\scriptsize},
    dist/.style={-{Latex[length=2mm]}, line width=0.8pt, black!85},
    nnline/.style={line width=0.7pt, black!65},
    smalllabel/.style={font=\scriptsize, fill=white, inner sep=1pt},
    neighbg/.style={
        fill=orange!35,
        fill opacity=0.16,
        draw=orange!70!black,
        draw opacity=0.75,
        line width=0.7pt,
        pattern={Lines[angle=45,distance=4pt,line width=0.25pt]},
        pattern color=orange!70!black
    }
]

\def\Lcx{0.95}
\def\Lcy{1.60}
\def\Rcx{3.45}
\def\Rcy{1.60}
\def\Ix{2.20}
\def\Iy{1.28}

\begin{scope}
\node[title] at (2.20,2.95) {(a) Centroid distances};

\draw[topic] (\Lcx,\Lcy) circle (0.55);
\node[pt] at (0.70,1.85) {};
\node[pt] at (0.90,1.96) {};
\node[pt] at (1.18,1.82) {};
\node[pt] at (0.72,1.46) {};
\node[pt] at (1.14,1.36) {};
\node[pt] at (1.30,1.56) {};
\node[centroidmark] at (\Lcx,\Lcy) {$\times$};

\draw[topic] (\Rcx,\Rcy) circle (0.55);
\node[pt] at (3.18,1.84) {};
\node[pt] at (3.42,1.94) {};
\node[pt] at (3.68,1.78) {};
\node[pt] at (3.12,1.50) {};
\node[pt] at (3.45,1.32) {};
\node[pt] at (3.72,1.52) {};
\node[centroidmark] at (\Rcx,\Rcy) {$\times$};

\node[pt] at (2.58,1.93) {};
\node[pt] at (2.58,1.08) {};
\node[pt] at (1.46,0.90) {};

\node[pt] (iA) at (\Ix,\Iy) {};
\node[anchor=north] at (\Ix,\Iy-0.07) {$i$};

\draw[dist] (iA) -- (\Lcx,\Lcy)
    node[pos=0.46, below=2pt, fill=white, inner sep=1pt] {$d_1$};

\draw[dist] (iA) -- (\Rcx,\Rcy)
    node[pos=0.54, below=2pt, fill=white, inner sep=1pt] {$d_2$};

\end{scope}

\draw[black!30, line width=0.5pt] (4.55,0.25) -- (4.55,2.80);

\begin{scope}[shift={(5.15,0)}]
\node[title] at (2.20,2.95) {(b) Local density};

\draw[topic] (\Lcx,\Lcy) circle (0.55);
\node[pt] at (0.70,1.85) {};
\node[pt] at (0.90,1.96) {};
\node[pt] (l1) at (1.18,1.82) {};
\node[pt] at (0.72,1.46) {};
\node[pt] (l3) at (1.14,1.36) {};
\node[pt] (l2) at (1.30,1.56) {};
\node[centroidmark] at (\Lcx,\Lcy) {$\times$};

\draw[topic] (\Rcx,\Rcy) circle (0.55);
\node[pt] at (3.18,1.84) {};
\node[pt] at (3.42,1.94) {};
\node[pt] at (3.68,1.78) {};
\node[pt] (r1) at (3.12,1.50) {};
\node[pt] at (3.45,1.32) {};
\node[pt] at (3.72,1.52) {};
\node[centroidmark] at (\Rcx,\Rcy) {$\times$};

\node[pt] (oTR) at (2.58,1.93) {};
\node[pt] (oBR) at (2.58,1.08) {};
\node[pt] (oBL) at (1.46,0.90) {};

\node[pt] (iB) at (\Ix,\Iy) {};
\node[anchor=north] at (\Ix,\Iy-0.07) {$i$};

\begin{pgfonlayer}{background}
\filldraw[neighbg]
    (1.03,1.26)
    .. controls (1.02,1.58) and (1.10,1.86) .. (1.30,1.96)
    .. controls (1.68,2.15) and (2.22,2.18) .. (2.70,2.05)
    .. controls (3.02,1.96) and (3.24,1.76) .. (3.22,1.48)
    .. controls (3.20,1.24) and (2.98,1.02) .. (2.64,0.90)
    .. controls (2.28,0.76) and (1.84,0.74) .. (1.46,0.82)
    .. controls (1.18,0.90) and (1.04,1.04) .. (1.03,1.26)
    -- cycle;
\end{pgfonlayer}

\draw[nnline] (iB) -- (l1);
\draw[nnline] (iB) -- (l2);
\draw[nnline] (iB) -- (l3);
\draw[nnline] (iB) -- (r1);
\draw[nnline] (iB) -- (oTR);
\draw[nnline] (iB) -- (oBR);
\draw[nnline] (iB) -- (oBL);

\node[smalllabel] at (2.22,0.44) {$\bar d_i^{\,7\mathrm{NN}}$};

\end{scope}
\end{tikzpicture}
}
\caption{Geometric predictors at publication time \(\Ta\): (a) distances from outlier \(i\) to the nearest topic centroids and (b) an illustrative local-neighborhood configuration.}
\label{fig:geometric_features}
\end{figure}
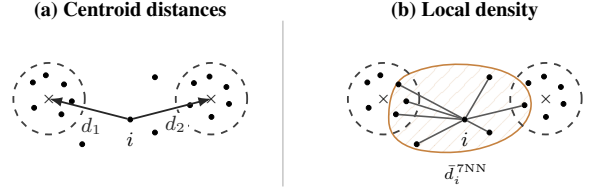

We first measure proximity to existing topics. For each cluster \(c \in C_{\Ta}\), we compute its centroid \(\mu_c\) and the Euclidean distance from \(x_i\) to \(\mu_c\). The nearest and second-nearest distances, \(d_{1i}\) and \(d_{2i}\), measure how far the article lies from the topic structure at \(\Ta\). Their difference, \(\Delta_i=d_{2i}-d_{1i}\), defines the centroid margin: large values indicate proximity to one dominant topic, while small values indicate comparable proximity to multiple topics. Figure~\ref{fig:geometric_features}(a) illustrates this geometry.

Centroid distances treat all clusters as equally compact, even though some topics are more dispersed than others. We therefore compute a diagonal Mahalanobis distance to each existing topic cluster containing at least five articles, and retain the minimum value
\[
d^{\,\mathrm{Mah}}_i =
\min_{\substack{c \in C_{\Ta}\\ |c|\geq 5}}
\sum_j
\frac{(x_{ij}-\mu_{cj})^2}{\sigma_{cj}^2+\epsilon},
\]
where \(\sigma_{cj}^2\) is the coordinate-wise variance of cluster \(c\), and \(\epsilon\) prevents division by zero. This feature measures whether \(x_i\) is atypical relative to the dispersion of the nearest plausible topic, rather than only distant in Euclidean space.

We then measure local density around the article. Using only articles observed in the cumulative snapshot at \(\Ta\), we compute the mean Euclidean distance \(\bar d^{\,\mathrm{kNN}}_i\) from \(x_i\) to up to 20 nearest neighbors, depending on the number of articles available in the snapshot; see Figure~\ref{fig:geometric_features}(b). We also compute the standard deviation of these distances to capture neighborhood heterogeneity and distinguish uniformly sparse regions from mixed neighborhoods containing both close and distant neighbors.

A further group of features captures proximity to other outliers. Let \(O(\Ta)\) denote the set of articles classified as outliers in the cumulative snapshot at \(\Ta\). We record whether this set is non-empty, its size, and the mean Euclidean distance from \(x_i\) to up to ten nearest articles in \(O(\Ta)\). These features distinguish outliers that are isolated in the embedding space from outliers that lie close to other publication-time outliers.

Finally, we include the \textit{outlierness score} at $T_A$, computed from the GLOSH score returned by HDBSCAN \cite{campello2015hierarchical,mcinnes2017hdbscan}. It measures how strongly an article is separated from dense regions in the HDBSCAN hierarchy. Higher values indicate that the article is weakly connected to nearby clusters. Unlike centroid-distance features, it captures local density structure rather than distance to topic centers.

Because each embedding model induces its own UMAP space, raw distance values are not directly comparable across models. We therefore convert each distance-based geometric feature to a within-model percentile rank. For feature \(g\), article \(i\), and embedding model \(m\), let \(g_{im}(\Ta)\) be the publication-time value and let \(n_m(\Ta)\) be the number of rows in the corresponding model-specific feature table. We compute
\[
p^{(g)}_{im}(\Ta) =
\frac{\operatorname{rank}_{m,\Ta}(g_{im}(\Ta))}{n_m(\Ta)+1},
\]
using average ranks for ties. This yields percentile-ranked features in $(0,1)$,
where larger values indicate larger within-model distances.

Article-level geometric predictors are obtained by aggregating model-specific feature values for each article. For each geometric feature, we compute the mean, median, and standard deviation across embedding models to capture average position, central tendency, and cross-model variation.

\subsection{Textual Features}
\label{sec:predictor_families_text}

Textual features measure whether anticipatory outliers are distinctive in content or form. They are computed once per article from its textual content.

We first compute tone features. Subjectivity is obtained with the French version of TextBlob \cite{loria2018textblob}. Neutrality is derived from the French VADER compound score \cite{hutto2014vader}. Articles with strong positive or negative polarity receive lower neutrality scores, whereas articles with weak affective polarity receive higher scores.

We then compute length and surface-complexity features. Character and word count measure the amount of textual information available. Average sentence length in words, average word length in characters, total syllables, and average syllables per word describe the surface complexity and density of this information.

Finally, we extract named entities with the French \texttt{spaCy} pipeline \cite{honnibal2020spacy}. The total number of entities measures the degree of entity anchoring in the article. The number of distinct entities captures how varied this anchoring is. We also retain counts for persons, organizations, locations, and miscellaneous entities, which indicate whether the article is tied to specific actors, institutions, places, or other named references. Such anchoring may precede cluster formation when later articles repeatedly refer to the same entities.

\subsection{Social Features}

Social features capture early circulation on X using URL-sharing events observed no later than the publication-time cumulative snapshot at \(\Ta\). If an article has no observed social trace by \(\Ta\), all social variables are set to zero, so that the absence of early circulation is encoded explicitly.

We first measure direct sharing activity. For each article URL, we record the number of distinct users who shared it, the median follower count, tweet count, and listed count of these users to capture the breadth and typical visibility of early circulation.

We then encode audience overlap through a weighted article--article co-sharing graph, obtained as a one-mode projection of the underlying user--article bipartite graph. Nodes correspond to article URLs observed by \(\Ta\). Two URLs are connected when at least one user shared both URLs by \(\Ta\), and the edge weight is the number of shared users. The derived metrics capture the extent to which new outlier articles circulate through overlapping audiences before a topic is semantically consolidated \cite{granovetter1973strength,ugander2012structural}.

From this graph we retain three measures. The weighted clustering coefficient \cite{onnela2005intensity} indicates whether the article belongs to a locally coherent co-sharing neighborhood. The Louvain community size \cite{blondel2008fast} measures the scale of the audience group in which the article circulates; the partition is computed with a fixed random seed, so that community assignments are reproducible. The bridge ratio measures the share of an article's co-sharing weight that connects outside its own community.

\section{Experimental Setup}
\label{sec:methodology}

\subsection{Datasets}
\label{sec:experiments:datasets}

We evaluate the prediction task on two French news corpora.
\textsc{HydroNewsFr}, described in \citet{zve-etal-2026-noise}, focuses on hydrogen-energy narratives. We curated \textsc{ClimateNewsFr}, a climate-change corpus collected with the same pipeline. It combines two complementary streams: news articles retrieved from Google News results using the GNews Python library, and X posts linking to news articles, collected through the official X API. The corpus was collected using the French keyword ``changement climatique'' (``climate change'') and includes the available title and lead paragraph.

\begin{table}[H]
\centering
\small
\setlength{\tabcolsep}{1.8pt}
\begin{tabular}{llrr}
\toprule
Corpus & Period & Articles & X posts \\
\midrule
\textsc{HydroNewsFr}   & 20 Mar--8 Jun 2025 & 1{,}616 & 1{,}533 \\
\textsc{ClimateNewsFr} & 27 Mar--27 May 2025 & 2{,}445 & 1{,}046 \\
\bottomrule
\end{tabular}
\caption{Summary of the two main corpora.}
\label{tab:corpora}
\end{table}

Both corpora are temporally ordered, multi-source news streams with limited
coverage gaps. This property is important for the present task, since the
transition of publication-time outliers into later topics can only be evaluated
reliably when the underlying stream is dense enough to support temporal
continuity. \textsc{ClimateNewsFr} is broader and more heterogeneous, covering
political, social, and environmental aspects of climate change, whereas
\textsc{HydroNewsFr} is centered on hydrogen-related industrial, technological,
and policy developments.

\subsection{Labeling-Pipeline Configuration}
\label{sec:clustering_setup}

In all experiments, we use an ensemble of 11 French-specific and multilingual embedding models. The full list is given in Appendix~\ref{app:embedding_models}. For each model, we concatenate each article's title and lead paragraph, embed the
resulting text, project the embeddings to 20 dimensions with UMAP, cluster them daily with HDBSCAN, and align clusters across time using centroid matching with threshold \(\theta_{\mathrm{align}} = 0.30\). This configuration follows the setting reported by \citet{zve-etal-2026-noise}, which was found to yield strong inter-model agreement on the \(\TOA\) versus non-\(\TOA\) distinction. We adapt their publicly available code to derive trajectory labels.\footnote{Code: \url{https://github.com/evangeliazve/aacl-public}} We keep the HDBSCAN library defaults, \feat{min\_cluster\_size=5} and \feat{min\_samples=5}.

\subsection{Supervised Learning and Evaluation}
\label{sec:ml_eval}

The supervised dataset pairs the consensus labels from Section~\ref{sec:labeling}
with the article-level publication-time predictors from
Section~\ref{sec:predictors}. Articles with uncertain consensus labels are
excluded from supervised training and evaluation. We also discard the first five days of each corpus as a warm-up window before training. We compare five classifiers:
logistic regression with \(\ell_2\) regularization
\cite{hoerl1970ridge}, a linear support vector classifier
\cite{cortes1995support}, a decision tree
\cite{breiman1984classification}, a random forest
\cite{breiman2001random}, and XGBoost \cite{chen2016xgboost}.

Because anticipatory outliers are the rarer class, we use \(F_1\) as the
primary metric and report recall and precision as complementary measures.
We estimate performance with two protocols. The main protocol is 5-fold article-level cross-validation, which measures how much information publication-time features carry about later trajectories. We complement it with a chronological forward-chaining protocol, in which each article is predicted by a model trained only on strictly earlier articles, and which therefore approximates deployment conditions. We also report a constant-positive baseline that predicts \(\TOA\) for all eligible
articles. The complete setup is provided in Appendix~\ref{app:evaluation_protocol}.

\section{Results}
\label{sec_results}

We examine whether anticipatory outliers can be predicted from information
available at publication time. The analysis proceeds in three steps. We first
vary the consensus threshold used to filter and label publication-time outliers,
assessing how prediction changes as supervision is supported by stronger
agreement across embedding models. We then compare multiple classifier families under
the selected consensus settings, to determine whether the signal is specific to one
learning algorithm or remains stable across modeling choices. Finally, we evaluate prediction chronologically.

\subsection{Effect of Agreement Thresholds}
\label{sec:results:agreement}

Table~\ref{tab:threshold_tradeoff_main} reports the coverage--performance tradeoff obtained by varying the embedding-model consensus threshold \(k\), using XGBoost at \(\Ta\). As introduced in Section~\ref{sec:labeling:consensus}, \(k\) specifies how many embedding models must agree that an article is a publication-time outlier and, for positive labels, that it follows an anticipatory trajectory (\TOA).

Increasing \(k\) makes the supervision set more selective. Fewer articles are retained, but their labels are supported by stronger inter-model agreement. At \(k=1\), the model already reaches \(F_1=0.765\) on \textsc{HydroNewsFr}
and \(F_1=0.778\) on \textsc{ClimateNewsFr}. Performance tends to improve under stricter agreement, consistent with reduced label noise, but high thresholds retain few positives and show greater fold-level variability. For the detailed classifier and interpretability analyses, we use \(k=4\) for
\textsc{HydroNewsFr} (83 positives, 254 negatives) and \(k=6\) for \textsc{ClimateNewsFr} (68 positives, 330 negatives). These are the largest
agreement thresholds before the positive class becomes very small and fold-level estimates become visibly less stable, as shown in
Table~\ref{tab:threshold_tradeoff_main}.

\begin{table*}[!t]
\centering
\small
\setlength{\tabcolsep}{6pt}
\renewcommand{\arraystretch}{1.08}
\begin{tabular}{
  c
  S[table-format=4.0] S[table-format=3.0] S[table-format=4.0]
  r@{\,\(\pm\)\,}l
  @{}p{1.5em}@{}
  S[table-format=4.0] S[table-format=3.0] S[table-format=4.0]
  r@{\,\(\pm\)\,}l
}
\toprule
& \multicolumn{5}{c}{\textsc{HydroNewsFr}} &
& \multicolumn{5}{c}{\textsc{ClimateNewsFr}} \\
\cmidrule(lr){2-6}\cmidrule(lr){8-12}
\(k\) & {Retained} & {Positive} & {Negative} & \multicolumn{2}{c}{\(F_1 \pm\) std} &
      & {Retained} & {Positive} & {Negative} & \multicolumn{2}{c}{\(F_1 \pm\) std} \\
\midrule
1 & 1235 & 544 & 691 & 0.765 & 0.013 && 1941 & 875 & 1066 & 0.778 & 0.012 \\
2 &  773 & 276 & 497 & 0.811 & 0.056 && 1390 & 489 &  901 & 0.851 & 0.016 \\
3 &  508 & 150 & 358 & 0.895 & 0.022 && 1005 & 273 &  732 & 0.895 & 0.020 \\
4 &  337 &  83 & 254 & \textbf{0.912} & \textbf{0.069} &&  766 & 182 &  584 & 0.915 & 0.038 \\
5 &  219 &  39 & 180 & 0.860 & 0.116 &&  570 & 115 &  455 & 0.952 & 0.033 \\
6 &  145 &  21 & 124 & 0.910 & 0.092 &&  398 &  68 &  330 & \textbf{0.969} & \textbf{0.047} \\
7 &   81 &  13 &  68 & 0.917 & 0.144 &&  270 &  43 &  227 & 0.940 & 0.016 \\
8 &   40 &   6 &  34 & 0.800 & 0.400 &&  161 &  21 &  140 & 0.949 & 0.063 \\
\bottomrule
\end{tabular}
\caption{Effect of the agreement threshold \(k\) at \(\Ta\) with XGBoost. Scores are cross-validated means \(\pm\) standard deviations across folds. Bold marks the \(F_1\) scores of the selected \(k\) values used in the detailed analyses.}
\label{tab:threshold_tradeoff_main}
\end{table*}

\subsection{Prediction at Publication Time}
\label{sec:results:prediction}

Using the selected corpus-specific values of \(k\), we train and compare five classifier families on the corresponding article-level labeled subsets. Table~\ref{tab:perf_models_ta_main} reports cross-validated
performance at \(\Ta\).

All trained models substantially outperform the constant-positive baseline, indicating that publication-time features contain information about later integration into newly formed topics when trajectory labels are supported by
consistent evidence across embedding models. XGBoost obtains the highest \(F_1\)-score in both corpora: \(0.912{\pm}0.069\) in \textsc{HydroNewsFr} and \(0.969{\pm}0.047\) in \textsc{ClimateNewsFr}. Linear models are more recall-oriented, with Linear SVC and logistic regression reaching recall \(0.941{\pm}0.064\) in
\textsc{HydroNewsFr}, and logistic regression reaching recall \(0.985{\pm}0.031\) in \textsc{ClimateNewsFr}.

\begin{table}[t]
\centering
\scriptsize
\setlength{\tabcolsep}{2.2pt}
\renewcommand{\arraystretch}{1.08}
\begin{tabular}{lcccccc}
\toprule
& \multicolumn{3}{c}{\textsc{HydroNewsFr}}
& \multicolumn{3}{c}{\textsc{ClimateNewsFr}} \\
\cmidrule(lr){2-4}\cmidrule(lr){5-7}
Model & Precision & \(F_1\)-score & Recall & Precision & \(F_1\)-score & Recall \\
\midrule
Baseline
& \(0.246\) & \(0.395\) & \(1.000\)
& \(0.171\) & \(0.291\) & \(1.000\) \\
Linear SVC
& \(0.853\) & \(0.889\) & \(\mathbf{0.941}\)
& \(0.889\) & \(0.906\) & \(0.930\) \\
LogReg \(\ell_2\)
& \(0.838\) & \(0.882\) & \(\mathbf{0.941}\)
& \(0.895\) & \(0.936\) & \(\mathbf{0.985}\) \\
Decision Tree
& \(0.803\) & \(0.835\) & \(0.891\)
& \(0.867\) & \(0.898\) & \(0.937\) \\
Random Forest
& \(0.867\) & \(0.885\) & \(0.917\)
& \(0.905\) & \(0.935\) & \(0.969\) \\
XGBoost
& \(\mathbf{0.919}\) & \(\mathbf{0.912}\) & \(0.917\)
& \(\mathbf{0.970}\) & \(\mathbf{0.969}\) & \(0.969\) \\
\bottomrule
\end{tabular}
\caption{Article-level precision, \(F_1\)-score, and recall at
\(\Ta\) under selected consensus settings: \(k=4\) for
\textsc{HydroNewsFr} and \(k=6\) for \textsc{ClimateNewsFr}.
Results are averaged over five article-level CV folds.}
\label{tab:perf_models_ta_main}
\end{table}

\subsection{Chronological Evaluation}
\label{sec:results:forward}

Cross-validation lets a model see articles published after those it is tested on. To assess performance under deployment conditions, we additionally evaluate XGBoost with chronological forward chaining: at each cutoff \(t\), the model is trained on all labeled articles with \(\Ta \le t\) and tested on those published in the following \(W\) days, so that each evaluated article is predicted exactly once by a model that has seen only strictly earlier articles. Table~\ref{tab:forward_chaining_main} reports pooled \(F_1\) for \(W=7\) days; the full protocol and other window lengths are given in Appendix~\ref{app:forward_chaining}.

\begin{table}[t]
\centering
\small
\setlength{\tabcolsep}{3pt}
\renewcommand{\arraystretch}{1.10}
\begin{tabular}{lrrll}
\toprule
Corpus & Pos. & Neg. & \(F_1\) [95\% CI] & Base \\
\midrule
\textsc{HydroNewsFr}   & 42 & 211 & 0.762 [0.65, 0.86] & 0.285 \\
\textsc{ClimateNewsFr} & 26 & 316 & 0.800 [0.66, 0.91] & 0.141 \\
\bottomrule
\end{tabular}
\caption{Forward-chaining results at \(\Ta\) with XGBoost and \(W=7\)-day test windows, under the selected consensus settings. Cells report \(F_1\) with a 95\% percentile bootstrap interval and the constant-positive baseline.}
\label{tab:forward_chaining_main}
\end{table}

As expected, chronological scores are lower than cross-validated ones, since each model is trained on fewer articles and cannot exploit any later structure of the stream. Both corpora nevertheless remain well above the constant-positive baseline even at the lower confidence bound, and results are stable across window lengths on the longer extended corpus. The cross-validated estimates should thus be read as an upper bound on the information carried by publication-time features, and the chronological estimates as a more conservative indication of prospective performance.

\section{Interpretability Analysis}
\label{sec:interpretability}

We analyze the learned XGBoost models from the cross-validated setting of Section~\ref{sec:results:prediction} to identify which publication-time features contribute to \TOA\ predictions. We first compare feature families through ablation, then inspect global SHAP importance and local out-of-fold explanations.

\subsection{Ablation Study}
\label{sec:results:ablation}

Table~\ref{tab:ablation_xgb_ta} reports feature-family ablations for XGBoost at \(\Ta\). ``Only'' rows use a single feature family, whereas ``All w/o'' rows remove one family from the full feature set. To test whether the main contrasts are stable across folds, we compare paired fold-level \(F_1\) scores from the same five article-level cross-validation splits using paired \(t\)-tests, with Benjamini--Hochberg correction \citep{benjamini1995controlling}.

Geometry accounts for most of the predictive signal. Using geometry alone nearly matches the full model in both corpora, with no significant \(F_1\) decrease in \textsc{HydroNewsFr} (\(0.900\) versus \(0.912\)) and no decrease in \textsc{ClimateNewsFr} (\(0.969\) versus \(0.969\)). In contrast, removing geometry yields a large and significant drop, from \(0.912\) to \(0.385\) in \textsc{HydroNewsFr} and from \(0.969\) to \(0.226\) in \textsc{ClimateNewsFr}. The precision--recall columns show that this effect is not driven by only one side of the tradeoff, as removing geometry lowers both precision and recall in both corpora.

Text and social features are weak as standalone predictors in both corpora, and both perform significantly below the geometry-only model. Removing either family from the full model does not significantly reduce \(F_1\). This suggests that these features provide at most auxiliary, corpus-specific information. The partial exception is social-only prediction in \textsc{HydroNewsFr}, with moderate precision but very low recall. This pattern indicates that social features help identify a small subset of anticipatory cases, but do not characterize the class.

\begin{table}[t]
\centering
\scriptsize
\setlength{\tabcolsep}{2.2pt}
\renewcommand{\arraystretch}{1.15}
\begin{tabular}{lcccccc}
\toprule
& \multicolumn{3}{c}{\textsc{HydroNewsFr}}
& \multicolumn{3}{c}{\textsc{ClimateNewsFr}} \\
\cmidrule(lr){2-4}\cmidrule(lr){5-7}
Feature set & Precision & \(F_1\) & Recall & Precision & \(F_1\) & Recall \\
\midrule
All features
& \(0.919\) & \(0.912\) & \(0.917\)
& \(0.970\) & \(0.969\) & \(0.969\) \\
Only geometry
& \(0.907\) & \(0.900\) & \(0.905\)
& \(0.970\) & \(0.969\) & \(0.969\) \\
Only text
& \(0.322\) & \(0.272^{\ddagger}\) & \(0.242\)
& \(0.279\) & \(0.202^{\ddagger}\) & \(0.165\) \\
Only social
& \(0.707\) & \(0.349^{\ddagger}\) & \(0.239\)
& \(0.235\) & \(0.083^{\ddagger}\) & \(0.212\) \\
All w/o geometry
& \(0.406\) & \(0.385^{\dagger}\) & \(0.371\)
& \(0.285\) & \(0.226^{\dagger}\) & \(0.194\) \\
All w/o social
& \(0.930\) & \(0.917\) & \(0.917\)
& \(0.970\) & \(0.969\) & \(0.969\) \\
All w/o text
& \(0.919\) & \(0.905\) & \(0.904\)
& \(0.969\) & \(0.962\) & \(0.955\) \\
\bottomrule
\end{tabular}
\caption{Ablations for XGBoost at $T_A$. Results are 5-fold CV means. Symbols mark significant \(F_1\) drops at $\alpha = 0.05$: $\dagger$ vs. all features; $\ddagger$ vs. geometry only.}
\label{tab:ablation_xgb_ta}
\end{table}

\subsection{SHAP Analysis}
\label{sec:results:shap}

\subsubsection{Global Importance}
\label{sec:results:global-shap}

Table~\ref{tab:shap_ta_top10_main} reports the top global SHAP features for the XGBoost models, computed with TreeExplainer \citep{lundberg2020local}. Mean absolute SHAP values measure each feature's average contribution magnitude, while Spearman correlations between feature values and signed SHAP values indicate whether larger feature values push predictions toward or away from \(\TOA\).

\begin{table*}[!t]
\centering
\begin{adjustbox}{max width=0.99\textwidth}
\small
\begin{tabular}{l c c c l c c c}
\toprule
\multicolumn{4}{c}{\textsc{HydroNewsFr}} & \multicolumn{4}{c}{\textsc{ClimateNewsFr}} \\
\cmidrule(lr){1-4}\cmidrule(lr){5-8}
Feature & Cat. & Mean \( |\mathrm{SHAP}| \) & Spearman \(r\) &
Feature & Cat. & Mean \( |\mathrm{SHAP}| \) & Spearman \(r\) \\
\midrule
\feat{d2\_second\_centroid\_pct\_median} & G & 0.1018 & \(0.69^{***}\uparrow\) &
\feat{knn\_std\_k20\_pct\_median} & G & 0.0773 & \(0.28^{***}\uparrow\) \\
\feat{d1\_nearest\_centroid\_pct\_median} & G & 0.0544 & \(0.64^{***}\uparrow\) &
\feat{d1\_nearest\_centroid\_pct\_median} & G & 0.0619 & \(0.48^{***}\uparrow\) \\
\feat{d2\_second\_centroid\_pct\_mean} & G & 0.0509 & \(0.67^{***}\uparrow\) &
\feat{d2\_second\_centroid\_pct\_median} & G & 0.0495 & \(0.38^{***}\uparrow\) \\
\feat{d1\_nearest\_centroid\_pct\_mean} & G & 0.0420 & \(0.84^{***}\uparrow\) &
\feat{n\_recent\_outliers\_mean} & G & 0.0255 & \(-0.91^{***}\downarrow\) \\
\feat{knn\_mean\_k20\_pct\_median} & G & 0.0417 & \(0.48^{***}\uparrow\) &
\feat{d1\_nearest\_centroid\_pct\_mean} & G & 0.0145 & \(0.86^{***}\uparrow\) \\
\feat{knn\_mean\_k20\_pct\_mean} & G & 0.0264 & \(0.65^{***}\uparrow\) &
\feat{knn\_mean\_k20\_pct\_median} & G & 0.0139 & \(0.38^{***}\uparrow\) \\
\feat{n\_recent\_outliers\_median} & G & 0.0136 & \(-0.81^{***}\downarrow\) &
\feat{outlier\_score\_std} & G & 0.0103 & \(0.67^{***}\uparrow\) \\
\feat{outlier\_score\_median} & G & 0.0135 & \(0.89^{***}\uparrow\) &
\feat{outlier\_score\_mean} & G & 0.0074 & \(0.67^{***}\uparrow\) \\
\feat{knn\_std\_k20\_pct\_mean} & G & 0.0129 & \(0.36^{***}\uparrow\) &
\feat{d2\_second\_centroid\_pct\_mean} & G & 0.0067 & \(0.37^{***}\uparrow\) \\
\feat{n\_recent\_outliers\_mean} & G & 0.0119 & \(-0.87^{***}\downarrow\) &
\feat{n\_recent\_outliers\_std} & G & 0.0066 & \(-0.94^{***}\downarrow\) \\
\bottomrule
\end{tabular}
\end{adjustbox}
\caption{Top XGBoost features, ranked separately by mean absolute SHAP value. Arrows indicate the sign of the Spearman correlation between feature value and SHAP contribution. G = geometry, T = text, S = social. Significance coding: \(^{***}p<0.001\), \(^{**}p<0.01\), \(^{*}p<0.05\).}
\label{tab:shap_ta_top10_main}
\end{table*}

The top-ranked features are geometric in both corpora. In \textsc{HydroNewsFr}, the largest contribution comes from the median second-nearest-centroid distance \((|\mathrm{SHAP}|=0.1018)\), followed by the median nearest-centroid distance \((|\mathrm{SHAP}|=0.0544)\). In \textsc{ClimateNewsFr}, the strongest feature is the median standard deviation of 20-nearest-neighbor distances \((|\mathrm{SHAP}|=0.0773)\), followed by median nearest- and second-nearest-centroid distances. These features are positively correlated with signed SHAP values, indicating that articles farther from existing topic centroids, or located in sparser and more heterogeneous local neighborhoods, receive higher \TOA\ scores.

The number of outliers present in the snapshot at \(\Ta\) shows the opposite pattern. In both corpora, larger outlier counts push predictions away from \TOA, with strong negative correlations in \textsc{HydroNewsFr} \((r=-0.81\) and \(r=-0.87)\) and \textsc{ClimateNewsFr} \((r=-0.91\) and \(r=-0.94)\). Thus, the model does not treat the number of outliers present at \(\Ta\) as evidence of anticipation.
High \TOA\ scores are instead associated with articles that tend to be distant from existing topic centroids, lie in sparse and heterogeneous local neighborhoods, and receive higher HDBSCAN outlierness scores.

Textual and social features have smaller and more corpus-specific effects, as shown in Table~\ref{tab:nongeom_shap}. In \textsc{HydroNewsFr}, the strongest textual feature is miscellaneous entity count, while total and person-entity counts are negatively associated with \TOA. In \textsc{ClimateNewsFr}, subjectivity is the strongest textual feature. Social features contribute weakly in \textsc{HydroNewsFr}, led by weighted clustering in the article--article co-sharing graph, and have zero mean absolute SHAP value in \textsc{ClimateNewsFr}. These patterns support the ablation result, with non-geometric features providing auxiliary information rather than the main predictive signal.

\begin{table}[H]
\centering
\scriptsize
\setlength{\tabcolsep}{2.0pt}
\begin{tabular}{lcc}
\toprule
Feature & Mean \( |\mathrm{SHAP}| \) & Spearman \(r\) \\
\midrule
\multicolumn{3}{l}{\textsc{HydroNewsFr}} \\
\feat{ner\_misc} & 0.0113 & \(0.82^{***}\) \\
\feat{ner\_total\_ents} & 0.0097 & \(-0.79^{***}\) \\
\feat{ner\_person} & 0.0082 & \(-0.81^{***}\) \\
\feat{text\_subjectivity} & 0.0071 & \(0.82^{***}\) \\
\feat{media\_weighted\_clustering} & 0.0027 & \(0.53^{***}\) \\
\feat{media\_community\_size} & 0.0022 & \(0.49^{***}\) \\
\midrule
\multicolumn{3}{l}{\textsc{ClimateNewsFr}} \\
\feat{text\_subjectivity} & 0.0027 & \(0.73^{***}\) \\
\feat{ner\_misc} & 0.0025 & \(-0.81^{***}\) \\
\bottomrule
\end{tabular}
\caption{Main non-geometric SHAP features at \(\Ta\). Significance coding: \(^{***}p<0.001\).}
\label{tab:nongeom_shap}
\end{table}

\subsubsection{Local Explanations}
\label{sec:results:local}

We examine one true-positive out-of-fold example from each corpus, selected to illustrate substantively interpretable later topics. Each article is predicted by a fold-specific model that was not trained on it. Retrospective topic reconstruction is then used to interpret the topic the article later joined.

\textsc{HydroNewsFr.}
We consider the \emph{H2 Mobile} article on Germany's use of salt caverns for large-scale hydrogen storage.\footnote{\emph{H2 Mobile}, ``L'Allemagne mise sur ses cavernes salines pour le stockage massif d'hydrogène,'' (25 April 2025).} Its out-of-fold predicted probability at \(\Ta\) is 0.9983. It is labeled positive, with 5 votes across the embedding-model ensemble. In the retrospective \texttt{multilingual-e5-large} reconstruction, \(\Ta\) is 25 April 2025, while \(\Tt=\Ti=\) 9 May 2025. The article therefore appears before the topic becomes a cluster in this reconstruction. The later topic is centered on large-scale underground hydrogen storage in salt formations and includes articles on Storengy's storage projects, German cavern capacity, and pilot demonstrations in saline formations.\footnote{\emph{France 3 Régions}, ``Un site pilote européen de l'Ain valide le stockage de l'hydrogène dans du sel,'' (26 May 2025); \emph{EnergyNews}, ``Le stockage d'hydrogène en cavités salines : opportunités et défis pour l'industrie,'' (23 May 2025).}

The strongest positive contributions come from distances to existing topic structure, especially the median second-nearest-centroid distance \((+0.2034)\), the median nearest-centroid distance \((+0.1254)\), and the median \(20\)-nearest-neighbor distance \((+0.1087)\). The median outlier-score also contributes positively \((+0.0382)\). Together, these features place the article away from existing hydrogen-topic centers and in a sparse neighborhood. Non-geometric effects are small: subjectivity \((-0.0165)\) and the number of named entities \((-0.0094)\) push slightly away from \TOA, while the number of unique users sharing the article contributes positively \((+0.0033)\).

\textsc{ClimateNewsFr.}
We consider the \emph{CNRS} article on archaeological sites threatened by climate change.\footnote{\emph{CNRS}, ``Réchauffement climatique : l'archéologie face à la disparition des sites et des fouilles,'' (2 April 2025).} Its out-of-fold predicted probability at \(\Ta\) is 0.9978. It is labeled positive, with 10 votes. In the retrospective \texttt{mistral-embed} reconstruction, \(\Ta=\) 2 April 2025, while \(\Tt=\Ti=\) 17 April 2025, so this article is a \TOAfirst\ case. The article therefore appears before climate-related cultural heritage and archaeology becomes a cluster in this reconstruction. The later topic includes articles on Greek historical sites threatened by climate change and on the transformation of cultural heritage worldwide under climate change, all published later than the CNRS article.\footnote{\emph{Courrier International}, ``En Grèce, les sites historiques menacés par le changement climatique,'' (23 April 2025); \emph{La Croix}, ``Le changement climatique transforme notre patrimoine aux quatre coins de la planète,'' (17 April 2025).}

The largest positive contributions are the median nearest-centroid distance \((+0.2019)\), the median standard deviation of \(20\)-nearest-neighbor distances \((+0.1983)\), and the mean second-nearest-centroid distance \((+0.1038)\). The number of outliers at \Ta also supports the prediction: consistent with the negative global correlation reported above, the article was published when the number of outliers at \Ta was low, and this low count contributes positively \((+0.0832)\). HDBSCAN outlier-score variability contributes more modestly \((+0.0327)\). These features place the article away from existing climate topics, in a sparse and heterogeneous neighborhood.
Textual effects are much smaller: average sentence length is the main negative correction \((-0.0109)\), while subjectivity contributes weakly toward \TOA\ \((+0.0052)\). Social features do not materially affect this local prediction.

In both examples, \TOA\ predictions are driven by distance from existing topic centers combined with a non-uniform local neighborhood at \(\Ta\).

\section{Conclusion}

We introduced \emph{anticipatory outlier detection}, a prospective task that predicts if a publication-time outlier will later become an early member of a topic that has not yet formed. On two French news corpora, we show that this trajectory is predictable from publication-time information, especially under stricter inter-model consensus labels, and that the signal persists under a strictly chronological evaluation. Prediction is driven mainly by embedding-space geometry: anticipatory outliers lie far from existing topic centroids, in sparse and uneven neighborhoods. Textual and social variables provide weaker, more corpus-specific signals.

More broadly, this work raises the question of whether anticipatory outliers reflect a general property of dynamic embedding spaces beyond news, and even beyond text. Similar signals may arise in images, video, or audio, when initially isolated representations later become part of emerging semantic or stylistic structures.

\section*{Limitations}

Our methodology provides evidence that publication-time outliers can contain signals of later topic formation within an embedding-based framework. The evaluation relies on consensus labels obtained across embedding models, mitigating dependence on any single embedding space. Qualitative examples further indicate that several predicted outliers correspond to coherent later topics. Nevertheless, external validation remains necessary. A systematic human-annotation study would help assess how these signals are perceived in terms of novelty, relevance, and real-world significance.

The study relies on two French news corpora with complementary scopes. \textsc{HydroNewsFr} is an existing dataset focused on a specialized industrial and policy domain, while \textsc{ClimateNewsFr} is a curated dataset covering a broader and more heterogeneous public issue. Both corpora are temporally ordered news streams built from Google News results and X-sharing activity of news articles, with articles drawn from multiple media sources. This provides relatively dense time series, reduces timeline discontinuities, and limits dependence on a single source while keeping the language and collection pipeline controlled across settings. However, the comparison remains limited to two domains within a single language and does not cover other media ecosystems or longer temporal scales. Replication on larger multilingual corpora and longer observation windows would further assess the transferability of the approach.

The supervised labels depend on the retrospective trajectory-reconstruction pipeline. In the main setting, embeddings are projected with 20-dimensional UMAP, clustered with HDBSCAN, and aligned across cumulative snapshots using a fixed topic-alignment threshold \((\theta_{\mathrm{align}} = 0.30)\). This configuration follows prior trajectory-labeling work showing strong inter-model agreement for the anticipatory versus non-anticipatory distinction, with limited variation across UMAP dimensionalities \cite{zve-etal-2026-noise}. However, the present study adds a downstream prospective prediction task, so robustness at the labeling stage may not fully imply robustness of prediction scores. Future work should therefore test whether the observed stability across UMAP dimensionalities also holds for prediction, and should examine alternative dimensionality-reduction methods, clustering algorithms, and alignment thresholds.

Finally, social features are limited to observed X-sharing activity, which provides only a partial view of online circulation. This limited coverage may partly explain the weak contribution of social features in our models. Early diffusion may also occur on other social-media platforms or through channels not captured in the present data. Future work should extend the analysis to additional platforms and circulation channels.

\section*{Ethical Considerations}

In line with open-science principles, the code for reproducing the experiments is publicly available in a dedicated GitHub repository (Appendix~\ref{app:sup_material}). The embedding ensemble combines open-source and API-based models to assess robustness across model families used in research and applied settings. Where possible, we favor compact models to reduce unnecessary computational cost and environmental impact.

The study uses news articles retrieved from Google News results using the GNews Python library, and social-media sharing traces collected through the official X API. We do not redistribute raw text or raw social-media traces, since these may contain identifiable user activity and may be subject to publishers' rights, platform terms, personal-data restrictions, or copyright constraints. Data can be shared for research purposes only upon reasonable request.

The method is intended for research on weak-signal detection in dynamic text streams, not as a deployable monitoring or decision system. Its outputs are corpus-level signals, not evidence that an article is objectively important, novel, or likely to shape future events. Human annotation, external validation, and calibration would be required before any applied use. The main practical risk is a biased allocation of attention: some predictions may reflect corpus-specific patterns, and other relevant documents may be missed. Any adaptation should document data provenance, media coverage, language scope, and source selection, and account for uncertainty.

\section*{Acknowledgments}
EZ thanks Infopro Digital for granting her the time to pursue her PhD thesis alongside her work.

\bibliographystyle{acl_natbib}
\bibliography{latex/custom}

\begin{thebibliography}{39}
\providecommand{\natexlab}[1]{#1}

\bibitem[{Aggarwal(2017)}]{aggarwal2017outlier}
Charu~C. Aggarwal. 2017.
\newblock \href {https://doi.org/10.1007/978-3-319-47578-3} {\emph{Outlier
  Analysis}}, 2 edition.
\newblock Springer, Cham.

\bibitem[{Allan(2002)}]{allan2002topic}
James Allan, editor. 2002.
\newblock \href {https://doi.org/10.1007/978-1-4615-0933-2} {\emph{Topic
  Detection and Tracking: Event-Based Information Organization}}.
\newblock Kluwer Academic Publishers, Boston, MA.

\bibitem[{Ankerst et~al.(1999)Ankerst, Breunig, Kriegel, and
  Sander}]{ankerst1999optics}
Mihael Ankerst, Markus~M. Breunig, Hans-Peter Kriegel, and J{\"o}rg Sander.
  1999.
\newblock \href {https://doi.org/10.1145/304182.304187} {{OPTICS}: Ordering
  points to identify the clustering structure}.
\newblock In \emph{Proceedings of the 1999 ACM SIGMOD International Conference
  on Management of Data}, pages 49--60. ACM.

\bibitem[{Ansoff(1975)}]{ansoff1975weak}
H.~Igor Ansoff. 1975.
\newblock \href {https://doi.org/10.2307/41164635} {Managing strategic surprise
  by response to weak signals}.
\newblock \emph{California Management Review}, 18(2):21--33.

\bibitem[{Becker et~al.(2011)Becker, Naaman, and Gravano}]{becker2011beyond}
Hila Becker, Mor Naaman, and Luis Gravano. 2011.
\newblock \href {https://ojs.aaai.org/index.php/ICWSM/article/view/14116}
  {Beyond trending topics: Real-world event identification on twitter}.
\newblock In \emph{Proceedings of the Fifth International AAAI Conference on
  Weblogs and Social Media}, pages 438--441.

\bibitem[{Benjamini and Hochberg(1995)}]{benjamini1995controlling}
Yoav Benjamini and Yosef Hochberg. 1995.
\newblock \href {https://doi.org/10.1111/j.2517-6161.1995.tb02031.x}
  {Controlling the false discovery rate: A practical and powerful approach to
  multiple testing}.
\newblock \emph{Journal of the Royal Statistical Society: Series B
  (Methodological)}, 57(1):289--300.

\bibitem[{Blei and Lafferty(2006)}]{blei2006dynamic}
David~M. Blei and John~D. Lafferty. 2006.
\newblock \href {https://doi.org/10.1145/1143844.1143859} {Dynamic topic
  models}.
\newblock In \emph{Proceedings of the 23rd International Conference on Machine
  Learning}, pages 113--120. ACM.

\bibitem[{Blondel et~al.(2008)Blondel, Guillaume, Lambiotte, and
  Lefebvre}]{blondel2008fast}
Vincent~D. Blondel, Jean-Loup Guillaume, Renaud Lambiotte, and Etienne
  Lefebvre. 2008.
\newblock \href {https://doi.org/10.1088/1742-5468/2008/10/P10008} {Fast
  unfolding of communities in large networks}.
\newblock \emph{Journal of Statistical Mechanics: Theory and Experiment},
  2008(10):P10008.

\bibitem[{Boutaleb et~al.(2024)Boutaleb, Picault, and
  Grosjean}]{boutaleb2024bertrend}
Allaa Boutaleb, J{\'e}r{\^o}me Picault, and Guillaume Grosjean. 2024.
\newblock \href {https://aclanthology.org/2024.futured-1.1/} {{BERT}rend:
  Neural topic modeling for emerging trends detection}.
\newblock In \emph{Proceedings of the Workshop on the Future of Event
  Detection}, pages 1--17, Miami, Florida, USA. Association for Computational
  Linguistics.

\bibitem[{Breiman(2001)}]{breiman2001random}
Leo Breiman. 2001.
\newblock \href {https://doi.org/10.1023/A:1010933404324} {Random forests}.
\newblock \emph{Machine Learning}, 45(1):5--32.

\bibitem[{Breiman et~al.(1984)Breiman, Friedman, Olshen, and
  Stone}]{breiman1984classification}
Leo Breiman, Jerome~H. Friedman, Richard~A. Olshen, and Charles~J. Stone. 1984.
\newblock \emph{Classification and Regression Trees}.
\newblock Wadsworth, Belmont, CA.

\bibitem[{Campello et~al.(2015)Campello, Moulavi, Zimek, and
  Sander}]{campello2015hierarchical}
Ricardo J. G.~B. Campello, Davoud Moulavi, Arthur Zimek, and J{\"o}rg Sander.
  2015.
\newblock \href {https://doi.org/10.1145/2733381} {Hierarchical density
  estimates for data clustering, visualization, and outlier detection}.
\newblock \emph{ACM Transactions on Knowledge Discovery from Data},
  10(1):1--51.

\bibitem[{Chen and Guestrin(2016)}]{chen2016xgboost}
Tianqi Chen and Carlos Guestrin. 2016.
\newblock \href {https://doi.org/10.1145/2939672.2939785} {{XGBoost}: A
  scalable tree boosting system}.
\newblock In \emph{Proceedings of the 22nd ACM SIGKDD International Conference
  on Knowledge Discovery and Data Mining}, pages 785--794. ACM.

\bibitem[{Christophe et~al.(2021)Christophe, Velcin, Cugliari, Boumghar, and
  Suignard}]{christophe2021monitoring}
Cl{\'e}ment Christophe, Julien Velcin, Jairo Cugliari, Manel Boumghar, and
  Philippe Suignard. 2021.
\newblock \href {https://doi.org/10.18653/v1/2021.emnlp-main.76} {Monitoring
  geometrical properties of word embeddings for detecting the emergence of new
  topics}.
\newblock In \emph{Proceedings of the 2021 Conference on Empirical Methods in
  Natural Language Processing}, pages 994--1003. Association for Computational
  Linguistics.

\bibitem[{Cortes and Vapnik(1995)}]{cortes1995support}
Corinna Cortes and Vladimir Vapnik. 1995.
\newblock \href {https://doi.org/10.1007/BF00994018} {Support-vector networks}.
\newblock \emph{Machine Learning}, 20(3):273--297.

\bibitem[{Dieng et~al.(2020)Dieng, Ruiz, and Blei}]{dieng2020topic}
Adji~B. Dieng, Francisco J.~R. Ruiz, and David~M. Blei. 2020.
\newblock \href {https://doi.org/10.1162/tacl_a_00325} {Topic modeling in
  embedding spaces}.
\newblock \emph{Transactions of the Association for Computational Linguistics},
  8:439--453.

\bibitem[{Ebadi et~al.(2026)Ebadi, Auger, and Gauthier}]{ebadi2026wisdom}
Ashkan Ebadi, Alain Auger, and Yvan Gauthier. 2026.
\newblock \href {https://doi.org/10.1016/j.joi.2025.101759} {{WISDOM}: An
  {AI}-powered framework for emerging research detection using weak signal
  analysis and advanced topic modelling}.
\newblock \emph{Journal of Informetrics}, 20(1):101759.

\bibitem[{Granovetter(1973)}]{granovetter1973strength}
Mark~S. Granovetter. 1973.
\newblock \href {https://doi.org/10.1086/225469} {The strength of weak ties}.
\newblock \emph{American Journal of Sociology}, 78(6):1360--1380.

\bibitem[{Grootendorst(2022)}]{grootendorst2022bertopic}
Maarten Grootendorst. 2022.
\newblock \href {https://arxiv.org/abs/2203.05794} {{BERTopic}: Neural topic
  modeling with a class-based {TF-IDF} procedure}.
\newblock \emph{arXiv preprint arXiv:2203.05794}.

\bibitem[{Hiltunen(2008)}]{hiltunen2008future}
Elina Hiltunen. 2008.
\newblock \href {https://doi.org/10.1016/j.futures.2007.08.021} {The future
  sign and its three dimensions}.
\newblock \emph{Futures}, 40(3):247--260.

\bibitem[{Hoerl and Kennard(1970)}]{hoerl1970ridge}
Arthur~E. Hoerl and Robert~W. Kennard. 1970.
\newblock \href {https://doi.org/10.1080/00401706.1970.10488634} {Ridge
  regression: Biased estimation for nonorthogonal problems}.
\newblock \emph{Technometrics}, 12(1):55--67.

\bibitem[{Honnibal et~al.(2020)Honnibal, Montani, Van~Landeghem, and
  Boyd}]{honnibal2020spacy}
Matthew Honnibal, Ines Montani, Sofie Van~Landeghem, and Adriane Boyd. 2020.
\newblock \href {https://doi.org/10.5281/zenodo.1212303} {{spaCy}:
  Industrial-strength natural language processing in python}.

\bibitem[{Hutto and Gilbert(2014)}]{hutto2014vader}
C.~J. Hutto and Eric Gilbert. 2014.
\newblock \href {https://doi.org/10.1609/icwsm.v8i1.14550} {{VADER}: A
  parsimonious rule-based model for sentiment analysis of social media text}.
\newblock In \emph{Proceedings of the International AAAI Conference on Web and
  Social Media}, volume~8, pages 216--225.

\bibitem[{Kleinberg(2002)}]{kleinberg2002bursty}
Jon Kleinberg. 2002.
\newblock \href {https://doi.org/10.1145/775047.775061} {Bursty and
  hierarchical structure in streams}.
\newblock In \emph{Proceedings of the Eighth ACM SIGKDD International
  Conference on Knowledge Discovery and Data Mining}, pages 91--101. ACM.

\bibitem[{Kuhn(1955)}]{kuhn1955hungarian}
Harold~W. Kuhn. 1955.
\newblock \href {https://doi.org/10.1002/nav.3800020109} {The hungarian method
  for the assignment problem}.
\newblock \emph{Naval Research Logistics Quarterly}, 2(1--2):83--97.

\bibitem[{Loria(2018)}]{loria2018textblob}
Steven Loria. 2018.
\newblock {TextBlob} documentation.
\newblock \url{https://textblob.readthedocs.io/}.
\newblock Release 0.15.

\bibitem[{Lundberg et~al.(2020)Lundberg, Erion, Chen, DeGrave, Prutkin, Nair,
  Katz, Himmelfarb, Bansal, and Lee}]{lundberg2020local}
Scott~M. Lundberg, Gabriel Erion, Hugh Chen, Alex DeGrave, Jordan~M. Prutkin,
  Bala Nair, Ronit Katz, Jonathan Himmelfarb, Nisha Bansal, and Su-In Lee.
  2020.
\newblock \href {https://doi.org/10.1038/s42256-019-0138-9} {From local
  explanations to global understanding with explainable {AI} for trees}.
\newblock \emph{Nature Machine Intelligence}, 2(1):56--67.

\bibitem[{McInnes et~al.(2017)McInnes, Healy, and Astels}]{mcinnes2017hdbscan}
Leland McInnes, John Healy, and Steve Astels. 2017.
\newblock \href {https://doi.org/10.21105/joss.00205} {{hdbscan}: Hierarchical
  density based clustering}.
\newblock \emph{Journal of Open Source Software}, 2(11):205.

\bibitem[{McInnes et~al.(2018)McInnes, Healy, Saul, and
  Gro{\ss}berger}]{mcinnes2018umap}
Leland McInnes, John Healy, Nathaniel Saul, and Lukas Gro{\ss}berger. 2018.
\newblock \href {https://doi.org/10.21105/joss.00861} {{UMAP}: Uniform manifold
  approximation and projection}.
\newblock \emph{Journal of Open Source Software}, 3(29):861.

\bibitem[{Onnela et~al.(2005)Onnela, Saram{\"a}ki, Kert{\'e}sz, and
  Kaski}]{onnela2005intensity}
Jukka-Pekka Onnela, Jari Saram{\"a}ki, J{\'a}nos Kert{\'e}sz, and Kimmo Kaski.
  2005.
\newblock \href {https://doi.org/10.1103/PhysRevE.71.065103} {Intensity and
  coherence of motifs in weighted complex networks}.
\newblock \emph{Physical Review E}, 71(6):065103.

\bibitem[{Petrovi{\'c} et~al.(2010)Petrovi{\'c}, Osborne, and
  Lavrenko}]{petrovic2010streaming}
Sa{\v{s}}a Petrovi{\'c}, Miles Osborne, and Victor Lavrenko. 2010.
\newblock \href {https://aclanthology.org/N10-1021/} {Streaming first story
  detection with application to twitter}.
\newblock In \emph{Human Language Technologies: The 2010 Annual Conference of
  the North American Chapter of the Association for Computational Linguistics},
  pages 181--189. Association for Computational Linguistics.

\bibitem[{Pimentel et~al.(2014)Pimentel, Clifton, Clifton, and
  Tarassenko}]{pimentel2014review}
Marco A.~F. Pimentel, David~A. Clifton, Lei Clifton, and Lionel Tarassenko.
  2014.
\newblock \href {https://doi.org/10.1016/j.sigpro.2013.12.026} {A review of
  novelty detection}.
\newblock \emph{Signal Processing}, 99:215--249.

\bibitem[{Ratner et~al.(2017)Ratner, Bach, Ehrenberg, Fries, Wu, and
  R{\'e}}]{ratner2017snorkel}
Alexander Ratner, Stephen~H. Bach, Henry Ehrenberg, Jason Fries, Sen Wu, and
  Christopher R{\'e}. 2017.
\newblock \href {https://doi.org/10.14778/3157794.3157797} {Snorkel: Rapid
  training data creation with weak supervision}.
\newblock \emph{Proceedings of the VLDB Endowment}, 11(3):269--282.

\bibitem[{Snow et~al.(2008)Snow, O'Connor, Jurafsky, and Ng}]{snow2008cheap}
Rion Snow, Brendan O'Connor, Daniel Jurafsky, and Andrew~Y. Ng. 2008.
\newblock \href {https://aclanthology.org/D08-1027/} {Cheap and fast---but is
  it good? evaluating non-expert annotations for natural language tasks}.
\newblock In \emph{Proceedings of the 2008 Conference on Empirical Methods in
  Natural Language Processing}, pages 254--263. Association for Computational
  Linguistics.

\bibitem[{Strehl and Ghosh(2002)}]{strehl2002cluster}
Alexander Strehl and Joydeep Ghosh. 2002.
\newblock \href {https://www.jmlr.org/papers/v3/strehl02a.html} {Cluster
  ensembles---a knowledge reuse framework for combining multiple partitions}.
\newblock \emph{Journal of Machine Learning Research}, 3:583--617.

\bibitem[{Ugander et~al.(2012)Ugander, Backstrom, Marlow, and
  Kleinberg}]{ugander2012structural}
Johan Ugander, Lars Backstrom, Cameron Marlow, and Jon Kleinberg. 2012.
\newblock \href {https://doi.org/10.1073/pnas.1116502109} {Structural diversity
  in social contagion}.
\newblock \emph{Proceedings of the National Academy of Sciences},
  109(16):5962--5966.

\bibitem[{Wang and McCallum(2006)}]{wang2006topics}
Xuerui Wang and Andrew McCallum. 2006.
\newblock \href {https://doi.org/10.1145/1150402.1150450} {Topics over time: A
  non-markov continuous-time model of topical trends}.
\newblock In \emph{Proceedings of the 12th ACM SIGKDD International Conference
  on Knowledge Discovery and Data Mining}, pages 424--433. ACM.

\bibitem[{Zve et~al.(2026)Zve, Bourgne, Icard, and
  Ganascia}]{zve-etal-2026-noise}
Evangelia Zve, Gauvain Bourgne, Benjamin Icard, and Jean-Gabriel Ganascia.
  2026.
\newblock \href {https://doi.org/10.63317/5c6zvq4nbjdq} {From noise to signal:
  When outliers seed new topics}.
\newblock In \emph{Proceedings of the Fifteenth Language Resources and
  Evaluation Conference}, pages 7523--7533, Palma de Mallorca, Spain. ELRA
  Language Resource Association.

\bibitem[{Zve et~al.(2025)Zve, Icard, Breton, Sainero, Bourgne, and
  Ganascia}]{zve2025outliers}
Evangelia Zve, Benjamin Icard, Alice Breton, Lila Sainero, Gauvain Bourgne, and
  Jean-Gabriel Ganascia. 2025.
\newblock \href {https://aclanthology.org/2025.icnlsp-1.38/} {From outliers to
  topics in language models: Anticipating trends in news corpora}.
\newblock In \emph{Proceedings of the 8th International Conference on Natural
  Language and Speech Processing}, pages 385--398. Association for
  Computational Linguistics.

\end{thebibliography}
\appendix

\section{Supplementary Materials}
\label{app:sup_material}

We provide supplementary materials at:
\url{https://github.com/evangeliazve/aacl-public}. The repository includes the pipeline for applying the method to other corpora, together with the scripts used to reproduce the classification, ablation, SHAP, and forward-chaining results reported in the paper.

Data collection used GNews for Google News results\footnote{\url{https://github.com/ranahaani/GNews}} and the official X API\footnote{\url{https://developer.x.com/en/docs}}.

\section{Supplementary Experimental Details}
\label{app:experimental_details}
\subsection{Embedding Models}
\label{app:embedding_models}

Table~\ref{tab:embedding_models} lists the embedding models used for trajectory reconstruction.

\begin{table}[H]
\centering
\scriptsize
\begin{tabular}{lrrl}
\toprule
Model & Dim & Language & Access \\
\midrule
\href{https://huggingface.co/Lajavaness/sentence-camembert-base}{sentence-camembert-base} & 768 & French & Open \\
\href{https://huggingface.co/OrdalieTech/Solon-embeddings-large-0.1}{Solon-large-0.1} & 1024 & French & Open \\
\href{https://huggingface.co/sentence-transformers/paraphrase-multilingual-MiniLM-L12-v2}{paraphrase-MiniLM-L12-v2} & 384 & Multilingual & Open \\
\href{https://huggingface.co/sentence-transformers/paraphrase-multilingual-mpnet-base-v2}{paraphrase-mpnet-base-v2} & 768 & Multilingual & Open \\
\href{https://huggingface.co/sentence-transformers/LaBSE}{LaBSE} & 768 & Multilingual & Open \\
\href{https://huggingface.co/intfloat/multilingual-e5-large}{multilingual-e5-large} & 1024 & Multilingual & Open \\
\href{https://huggingface.co/Snowflake/snowflake-arctic-embed-l-v2.0}{arctic-embed-l-v2.0} & 1024 & Multilingual & Open \\
\href{https://huggingface.co/BAAI/bge-m3}{bge-m3} & 1024 & Multilingual & Open \\
\href{https://developers.openai.com/api/docs/models/text-embedding-3-small}{text-embedding-3-small} & 1536 & Multilingual & API \\
\href{https://ai.google.dev/gemini-api/docs/models/gemini-embedding-001}{gemini-embedding-001} & 3072 & Multilingual & API \\
\href{https://docs.mistral.ai/models/model-cards/mistral-embed-23-12}{mistral-embed} & 1024 & Multilingual & API \\
\bottomrule
\end{tabular}
\caption{Embedding models used in the experiments.}
\label{tab:embedding_models}
\end{table}

\subsection{Feature Glossary}
\label{app:glossary}

Table~\ref{tab:feature_glossary} presents the features proposed and discussed in this study, complementing Section~\ref{sec:predictors}. Geometric features are per-model quantities; the classifiers use their mean, median, and standard deviation across the eleven embedding models, aggregated as described in Section~\ref{sec:predictor_families_geometry}. Social and textual features are computed once per article; social features use the sharing events observed up to \(\Ta\), and textual features depend only on the article text.

\begin{table*}[t]
\centering
\scriptsize
\setlength{\tabcolsep}{3.5pt}
\renewcommand{\arraystretch}{1.08}
\begin{tabularx}{\textwidth}{p{3.0cm}p{4.0cm}X}
\toprule
Family & Feature & Definition / interpretation \\
\midrule
Geometric: topic distance & \feat{d1\_nearest\_centroid\_pct} & Within-model percentile rank of the Euclidean distance from the article embedding to the nearest non-noise topic centroid in the snapshot at \(\Ta\). \\
Geometric: topic distance & \feat{d2\_second\_centroid\_pct} & Within-model percentile rank of the Euclidean distance to the second-nearest non-noise topic centroid at \(\Ta\). \\
Geometric: topic distance & \feat{margin\_d2\_minus\_d1\_pct} & Within-model percentile rank of the margin between the second-nearest and nearest centroid distances at \(\Ta\). \\
Geometric: topic shape & \feat{mahal\_nearest\_pct} & Within-model percentile rank of the minimum diagonal Mahalanobis distance to an existing topic cluster at \(\Ta\). \\
Geometric: local density & \feat{knn\_mean\_k20\_pct} & Within-model percentile rank of the mean Euclidean distance to up to 20 nearest neighbors in the snapshot at \(\Ta\), fewer when the snapshot is smaller. \\
Geometric: local density & \feat{knn\_std\_k20\_pct} & Within-model percentile rank of the standard deviation of distances to these same neighbors at \(\Ta\). \\
Geometric: outlier neighborhood & \feat{outlier\_proto\_mean\_dist\_pct} & Within-model percentile rank of the mean distance to up to ten nearest articles classified as outliers in the snapshot at \(\Ta\). \\
Geometric: outlierness & \feat{outlier\_score} & HDBSCAN outlierness score at \(\Ta\), from the GLOSH score \cite{campello2015hierarchical,mcinnes2017hdbscan}. \\
Geometric: outlier pool & \feat{has\_recent\_outliers} & Indicator that at least one article is classified as an outlier in the snapshot at \(\Ta\). \\
Geometric: outlier pool & \feat{n\_recent\_outliers} & Number of articles classified as outliers in the snapshot at \(\Ta\). \\
Social diffusion & \feat{soc\_unique\_users} & Number of distinct X users who shared the article URL up to \(\Ta\). \\
Social diffusion & \feat{soc\_median\_user\_public\_metrics\_followers\_count} & Median follower count of the X users who shared the article URL up to \(\Ta\). \\
Social diffusion & \feat{soc\_median\_user\_public\_metrics\_tweet\_count} & Median lifetime tweet count of the X users who shared the article URL up to \(\Ta\). \\
Social diffusion & \feat{soc\_median\_user\_public\_metrics\_listed\_count} & Median listed count of the X users who shared the article URL up to \(\Ta\). \\
Media co-sharing graph & \feat{media\_weighted\_clustering} & Weighted clustering coefficient \cite{onnela2005intensity} of the article URL node in the co-sharing graph built from shares observed up to \(\Ta\). \\
Media co-sharing graph & \feat{media\_bridge\_ratio} & Share of the article node's co-sharing weight that connects outside its Louvain community \cite{blondel2008fast} in the co-sharing graph at \(\Ta\). \\
Media co-sharing graph & \feat{media\_community\_size} & Size of the Louvain community \cite{blondel2008fast} containing the article URL node in the co-sharing graph at \(\Ta\). \\
Text style & \feat{text\_subjectivity} & Subjectivity of the article text, from the French version of TextBlob \cite{loria2018textblob}. \\
Text style & \feat{text\_neutrality} & Neutrality of the article text, from the French VADER compound polarity magnitude \cite{hutto2014vader}. \\
Readability / length & \feat{avg\_sentence\_len\_words} & Average sentence length in words. \\
Readability / length & \feat{avg\_word\_len\_chars} & Average word length in characters. \\
Readability / length & \feat{total\_syllables} & Approximate total number of syllables in the article text, counted with a rule-based French heuristic. \\
Readability / length & \feat{avg\_syllables\_per\_word} & Approximate average number of syllables per word. \\
Readability / length & \feat{len\_chars} & Total number of characters in the article text. \\
Readability / length & \feat{len\_words} & Total number of word tokens in the article text. \\
Named entities & \feat{ner\_total\_ents} & Total number of named entities detected with the French spaCy pipeline \cite{honnibal2020spacy}. \\
Named entities & \feat{ner\_distinct\_ents} & Number of distinct named-entity strings detected with the same pipeline \cite{honnibal2020spacy}. \\
Named entities & \feat{ner\_person} & Count of person entities detected with the same pipeline \cite{honnibal2020spacy}. \\
Named entities & \feat{ner\_org} & Count of organization entities detected with the same pipeline \cite{honnibal2020spacy}. \\
Named entities & \feat{ner\_loc} & Count of location entities detected with the same pipeline \cite{honnibal2020spacy}. \\
Named entities & \feat{ner\_misc} & Count of named entities not assigned to person, organization, or location categories \cite{honnibal2020spacy}. \\
\bottomrule
\end{tabularx}
\caption{Glossary of predictors used in the supervised models.}
\label{tab:feature_glossary}
\end{table*}

\subsection{Evaluation Protocol}
\label{app:evaluation_protocol}

\paragraph{Warm-up period.}
Before training, we use the first five daily snapshots of each corpus as a
warm-up period, so that the cumulative clustering and trajectory-reconstruction
pipeline has sufficient prior context.

\paragraph{Cross-validation.}
Performance is estimated with 5-fold article-level cross-validation with approximately 80/20 train--test split in each fold, with four folds used for
training and the remaining fold held out for evaluation. Folds are constructed
at the article level, so that all records associated with the same article are
assigned to the same fold and no article can appear in both the training and test
sets.

\paragraph{Preprocessing.}
All preprocessing is fit within each training fold. Missing values are
median-imputed using the training split only. Logistic regression and Linear SVC
additionally use standardization fit on the training split. Tree-based models
are trained on imputed but unscaled features.

\paragraph{Class imbalance and baseline.}
For logistic regression, Linear SVC, decision tree, and random forest, we use
\feat{class\_weight=balanced}. For XGBoost, \feat{scale\_pos\_weight} is set once per corpus and consensus setting to
\(N_-/\max(N_+,1)\), where \(N_+\) and \(N_-\) are the positive and negative counts in the selected labeled subset for that setting.

\paragraph{SHAP Analysis.}
For the XGBoost interpretability analysis, we compute probability-scale SHAP values with TreeExplainer \citep{lundberg2020local}.

\subsection{Classifier Hyperparameters}
\label{app:hparams}

Table~\ref{tab:model_hyperparameters} details the fixed parameters of all classifiers used in the experiments.

\begin{table*}[t]
\centering
\footnotesize
\setlength{\tabcolsep}{2pt}
\renewcommand{\arraystretch}{1.15}
\begin{tabularx}{0.92\textwidth}{@{}p{0.13\textwidth}>{\raggedright\arraybackslash}X@{}}
\toprule
Model & Hyperparameters \\
\midrule
Random Forest &
\feat{n\_estimators=600}, \feat{max\_depth=8},
\feat{min\_samples\_split=20}, \feat{min\_samples\_leaf=10},
\feat{max\_features=sqrt}, \feat{class\_weight=balanced} \\[2pt]

LogReg \(\ell_2\) &
\feat{penalty=l2}, \feat{solver=liblinear},
\feat{max\_iter=2000}, \feat{class\_weight=balanced} \\[2pt]

Linear SVC &
\feat{max\_iter=5000}, \feat{class\_weight=balanced} \\[2pt]

Decision Tree &
\feat{max\_depth=6}, \feat{min\_samples\_leaf=10},
\feat{class\_weight=balanced} \\[2pt]

XGBoost &
\feat{n\_estimators=300}, \feat{max\_depth=4},
\feat{learning\_rate=0.05}, \feat{subsample=0.8},
\feat{colsample\_bytree=0.8}, \feat{objective=binary:logistic},
\feat{eval\_metric=logloss},
\feat{scale\_pos\_weight}=\(N_{-}/\max(N_{+},1)\) \\
\bottomrule
\end{tabularx}
\caption{Classifier hyperparameters used in the experiments. We use \feat{random\_state=42} where applicable. For XGBoost, \(N_+\) and \(N_-\) denote the positive and negative counts in the selected labeled subset for the corresponding corpus--consensus setting.}
\label{tab:model_hyperparameters}
\end{table*}

\section{Supplementary Classification Results}
\label{app:additional_results}

\subsection{Full Results at Selected \(k\)}
\label{app:full_results_selected_k}

Table~\ref{tab:full_results_selected_k} reports precision, recall, and \(F_1\)-score with standard deviations across folds for the selected consensus settings, complementing Table~\ref{tab:perf_models_ta_main}.

\begin{table*}[t]
\centering
\small
\setlength{\tabcolsep}{4pt}
\renewcommand{\arraystretch}{1.15}
\begin{tabular}{lcccccc}
\toprule
& \multicolumn{3}{c}{\textsc{HydroNewsFr}} & \multicolumn{3}{c}{\textsc{ClimateNewsFr}} \\
\cmidrule(lr){2-4}\cmidrule(lr){5-7}
Model & Precision & Recall & \(F_1\) & Precision & Recall & \(F_1\) \\
\midrule
Baseline      & 0.246{\(\pm\)}0.011 & 1.000{\(\pm\)}0.000 & 0.395{\(\pm\)}0.014 & 0.171{\(\pm\)}0.017 & 1.000{\(\pm\)}0.000 & 0.291{\(\pm\)}0.025 \\
Linear SVC    & 0.853{\(\pm\)}0.095 & 0.941{\(\pm\)}0.064 & 0.889{\(\pm\)}0.056 & 0.889{\(\pm\)}0.083 & 0.930{\(\pm\)}0.063 & 0.906{\(\pm\)}0.051 \\
LogReg \(\ell_2\) & 0.838{\(\pm\)}0.073 & 0.941{\(\pm\)}0.064 & 0.882{\(\pm\)}0.040 & 0.895{\(\pm\)}0.069 & 0.985{\(\pm\)}0.031 & 0.936{\(\pm\)}0.044 \\
Decision Tree & 0.803{\(\pm\)}0.083 & 0.891{\(\pm\)}0.115 & 0.835{\(\pm\)}0.051 & 0.867{\(\pm\)}0.085 & 0.937{\(\pm\)}0.058 & 0.898{\(\pm\)}0.058 \\
Random Forest & 0.867{\(\pm\)}0.103 & 0.917{\(\pm\)}0.080 & 0.885{\(\pm\)}0.063 & 0.905{\(\pm\)}0.079 & 0.969{\(\pm\)}0.038 & 0.935{\(\pm\)}0.057 \\
XGBoost       & 0.919{\(\pm\)}0.112 & 0.917{\(\pm\)}0.080 & 0.912{\(\pm\)}0.069 & 0.970{\(\pm\)}0.037 & 0.969{\(\pm\)}0.062 & 0.969{\(\pm\)}0.047 \\
\bottomrule
\end{tabular}
\caption{Full article-level cross-validated performance at \(\Ta\), reported as mean \(\pm\) standard deviation across folds. Results use \(k=4\) for \textsc{HydroNewsFr} and \(k=6\) for \textsc{ClimateNewsFr}.}
\label{tab:full_results_selected_k}
\end{table*}

\subsection{Classifier Results Across Agreement Thresholds}
\label{app:threshold_classifier_results}

Table~\ref{tab:appendix_all_classifiers_thresholds}
reports classifier \(F_1\)-scores across different values of
\(k\) at \(T_A\). These results complement the main XGBoost
threshold analysis in Section~\ref{sec:results:agreement}, by showing that the coverage--confidence pattern is
stable across classifier families.

\begin{table*}[t]
\centering
\small
\setlength{\tabcolsep}{2pt}
\renewcommand{\arraystretch}{1.20}
\begin{tabular}{llrrrrrrrrr}
\toprule
Corpus & \(k\) & Retained & Positive & Negative & Baseline & Linear SVC & LogReg & Tree & Forest & XGBoost \\
\midrule
\multirow{8}{*}{\textsc{HydroNewsFr}}
& 1 & 1235 & 544 & 691 & 0.611 & 0.765 & 0.772 & 0.712 & 0.761 & 0.765 \\
& 2 &  773 & 276 & 497 & 0.525 & 0.815 & 0.813 & 0.743 & 0.811 & 0.811 \\
& 3 &  508 & 150 & 358 & 0.455 & 0.851 & 0.841 & 0.820 & 0.854 & 0.895 \\
& 4 &  337 &  83 & 254 & 0.395 & 0.889 & 0.882 & 0.835 & 0.885 & 0.912 \\
& 5 &  219 &  39 & 180 & 0.298 & 0.863 & 0.864 & 0.764 & 0.843 & 0.860 \\
& 6 &  145 &  21 & 124 & 0.248 & 0.878 & 0.838 & 0.894 & 0.937 & 0.910 \\
& 7 &   81 &  13 &  68 & 0.330 & 0.950 & 0.950 & 0.867 & 0.833 & 0.917 \\
& 8 &   40 &   6 &  34 & 0.258 & 1.000 & 1.000 & 0.700 & 0.900 & 0.800 \\
\midrule
\multirow{8}{*}{\textsc{ClimateNewsFr}}
& 1 & 1941 & 875 & 1066 & 0.621 & 0.779 & 0.778 & 0.755 & 0.780 & 0.778 \\
& 2 & 1390 & 489 &  901 & 0.519 & 0.827 & 0.827 & 0.769 & 0.823 & 0.851 \\
& 3 & 1005 & 273 &  732 & 0.427 & 0.873 & 0.887 & 0.843 & 0.875 & 0.895 \\
& 4 &  766 & 182 &  584 & 0.384 & 0.905 & 0.903 & 0.860 & 0.910 & 0.915 \\
& 5 &  570 & 115 &  455 & 0.336 & 0.913 & 0.913 & 0.927 & 0.931 & 0.952 \\
& 6 &  398 &  68 &  330 & 0.291 & 0.906 & 0.936 & 0.898 & 0.935 & 0.969 \\
& 7 &  270 &  43 &  227 & 0.273 & 0.920 & 0.915 & 0.903 & 0.929 & 0.940 \\
& 8 &  161 &  21 &  140 & 0.228 & 0.929 & 0.927 & 0.962 & 0.985 & 0.949 \\
\bottomrule
\end{tabular}
\caption{Classifier results at \(\Ta\) across symmetric consensus thresholds \((k,k,0)\). Values are cross-validated mean \(F_1\) scores.}
\label{tab:appendix_all_classifiers_thresholds}
\end{table*}

\subsection{Fold-level Ablation Comparisons}
\label{app:ablation_significance}

Table~\ref{tab:ablation_significance} reports paired fold-level \(F_1\) comparisons for the main ablation contrasts in Table~\ref{tab:ablation_xgb_ta}. Tests are computed on the same five article-level cross-validation folds used to produce the ablation results. We report the mean paired difference \(\Delta F_1\), the paired \(t\)-test \(p\)-value, and the Benjamini--Hochberg corrected \(q\)-value \cite{benjamini1995controlling}.

\begin{table*}[t]
\centering
\small
\setlength{\tabcolsep}{5pt}
\renewcommand{\arraystretch}{1.20}
\begin{tabular}{llccc}
\toprule
Corpus & Comparison & \(\Delta F_1\) & \(p\) & \(q\) \\
\midrule
\textsc{HydroNewsFr}
& All vs. all w/o geometry
& \(+0.526\) & \(7.19{\times}10^{-4}\) & \(1.83{\times}10^{-3}\) \\
& All vs. only geometry
& \(+0.011\) & \(0.536\) & \(0.577\) \\
& All vs. all w/o social
& \(-0.006\) & \(0.374\) & \(0.436\) \\
& All vs. all w/o text
& \(+0.006\) & \(0.374\) & \(0.436\) \\
& Only geometry vs. only text
& \(+0.628\) & \(3.66{\times}10^{-4}\) & \(1.17{\times}10^{-3}\) \\
& Only geometry vs. only social
& \(+0.551\) & \(1.39{\times}10^{-4}\) & \(5.74{\times}10^{-4}\) \\
\midrule
\textsc{ClimateNewsFr}
& All vs. all w/o geometry
& \(+0.743\) & \(1.57{\times}10^{-4}\) & \(4.06{\times}10^{-4}\) \\
& All vs. only geometry
& \(0.000\) & -- & -- \\
& All vs. all w/o social
& \(0.000\) & -- & -- \\
& All vs. all w/o text
& \(+0.007\) & \(0.374\) & \(0.374\) \\
& Only geometry vs. only text
& \(+0.767\) & \(7.74{\times}10^{-5}\) & \(3.89{\times}10^{-4}\) \\
& Only geometry vs. only social
& \(+0.886\) & \(3.63{\times}10^{-4}\) & \(6.95{\times}10^{-4}\) \\
\bottomrule
\end{tabular}
\caption{Paired fold-level \(F_1\) comparisons for the selected ablation settings. Positive \(\Delta F_1\) means that the first condition outperforms the second. Dashes indicate identical fold-level \(F_1\) values or degenerate zero differences, for which the paired \(t\)-test is not defined.}
\label{tab:ablation_significance}
\end{table*}

\section{Robustness Check: Extended Hydrogen Corpus}
\label{app:extended}

We repeated the experiment on an extended hydrogen corpus spanning 1 January 2025 to 8 June 2025. This analysis is not part of the main evaluation because social features were not uniformly available over the full period. We use only geometric and textual features and interpret the results as supporting evidence under a longer observation window. Table \ref{tab:extended_hydro_symmetric_thresholds} shows the same pattern observed in the main experiments: stricter \(k\) retain fewer articles but yield stronger predictive performance.

\begin{table*}[t]
\centering
\small
\setlength{\tabcolsep}{4.0pt}
\renewcommand{\arraystretch}{0.98}
\begin{tabular}{rrrrrr}
\toprule
\(k\) & Retained & Positive & Negative & \(F_1\) & Recall \\
\midrule
1 & 2134 & 1056 & 1078 & 0.793 & 0.773 \\
2 & 1390 & 605  & 785  & 0.871 & 0.865 \\
3 & 965  & 387  & 578  & 0.922 & 0.931 \\
4 & 656  & 240  & 416  & 0.923 & 0.910 \\
5 & 465  & 163  & 302  & 0.942 & 0.939 \\
6 & 321  & 107  & 214  & 0.981 & 0.981 \\
7 & 204  & 71   & 133  & 0.981 & 0.988 \\
8 & 120  & 48   & 72   & 0.991 & 0.983 \\
\bottomrule
\end{tabular}
\caption{XGBoost results on the extended corpus.}
\label{tab:extended_hydro_symmetric_thresholds}
\end{table*}

\section{Forward-Chaining Evaluation}
\label{app:forward_chaining}

This appendix details the chronological evaluation summarized in Section~\ref{sec:results:forward}. At each cutoff \(t\), the model is trained on all articles with \(\Ta \le t\) and tested on those with \(\Ta \in (t, t+W]\); the cutoff then advances by \(W\), so each evaluated article is predicted exactly once, by a model trained only on strictly earlier articles. Origins are restricted to \(t \le \Ta^{\max} - W\), so the final partial window is not evaluated. The warm-up before the first origin follows an a-priori minimum-training-size rule, two weeks for the main corpora and four weeks for the extended corpus of Appendix~\ref{app:extended}. The decision threshold is fixed at 0.5 and XGBoost follows Appendix~\ref{app:hparams}, with \feat{scale\_pos\_weight} recomputed on each training set and median imputation fit on training articles only. Articles in the warm-up and in the final partial window are not evaluated, which is why the counts in Table~\ref{tab:forward_chaining} are smaller than in Table~\ref{tab:threshold_tradeoff_main}. Brackets give 95\% percentile bootstrap intervals on pooled \(F_1\) (2{,}000 article resamples). Results come from a single run in one pinned environment.

\begin{table*}[t]
\centering
\small
\setlength{\tabcolsep}{4pt}
\renewcommand{\arraystretch}{1.15}
\begin{tabular}{lrrlllll}
\toprule
Corpus & Pos. & Neg. & \(W\!=\!5\)d & \(W\!=\!7\)d & \(W\!=\!14\)d & \(W\!=\!21\)d & Base \\
\midrule
\textsc{HydroNewsFr}
& 42 & 211
& 0.776 [0.67, 0.86] & 0.762 [0.65, 0.86] & 0.780 [0.68, 0.87] & 0.696 [0.60, 0.79] & 0.285 \\
\textsc{ClimateNewsFr}
& 26 & 316
& 0.821 [0.70, 0.92] & 0.800 [0.66, 0.91] & 0.737 [0.59, 0.85] & 0.778 [0.64, 0.89] & 0.141 \\
\textsc{HydroNewsFr} Ext.
& 131 & 345
& 0.875 [0.83, 0.92] & 0.885 [0.84, 0.92] & 0.874 [0.83, 0.91] & 0.873 [0.83, 0.91] & 0.432 \\
\bottomrule
\end{tabular}
\caption{Forward-chaining results at \(\Ta\), for test windows of \(W\) days. Ext.\ denotes the extended hydrogen corpus of Appendix~\ref{app:extended}. Cells report pooled \(F_1\) with a 95\% percentile bootstrap interval; Base is the constant-positive baseline. Counts are the pooled evaluated articles at \(W\!=\!7\)d and vary slightly with \(W\) under the window-fit rule.}
\label{tab:forward_chaining}
\end{table*}

All three corpora exceed the constant-positive baseline even at the lower confidence bound. The extended corpus reaches the highest scores; it also covers a longer period and contains a higher proportion of positives. Under the window-fit rule, the extended corpus pools 129 positives and 290 negatives at \(W=14\)d; with an 8-week warm-up, \(F_1 = 0.851\) [0.79, 0.91]. Scores are stable across window lengths on the extended corpus and more variable on the two smaller ones, where each window contains few positives. Per-window tables, the full warm-up \(\times\) window grid, and per-subclass predicted-positive rates are available upon request.

\end{document}